%% file: main.tex
\documentclass[10pt,twocolumn,letterpaper]{article}

\usepackage[pagenumbers]{cvpr}

\input{preamble}

\definecolor{cvprblue}{rgb}{0.21,0.49,0.74}
\usepackage[pagebackref,breaklinks,colorlinks,allcolors=cvprblue]{hyperref}

\title{The Old Scene Lives in the Cache: Diagnosing and Shedding Stale History\\in Streaming Video Generation}

\input{authors}

\hypersetup{pdftitle={The Old Scene Lives in the Cache: Diagnosing and Shedding Stale History in Streaming Video Generation},pdfauthor={Hao Zhou, Ziyang Xu, Chenhan Jin, Hao Qin, Shengze Xu, Tieyong Zeng}}
\begin{document}
\maketitle
\input{fig_en/teaser}
\input{sec_en/0_abstract}
\input{sec_en/1_intro}
\input{sec_en/2_related}
\input{sec_en/3_prelim}
\input{sec_en/4_diagnosis}
\input{sec_en/5_method}
\input{sec_en/6_experiments}
\input{sec_en/7_conclusion}
\clearpage
\input{sec/8_llm_statement}
{
    \small

\input{main.bbl}
}
\clearpage
\input{supplement}

\end{document}

%% file: preamble.tex
\usepackage[T1]{fontenc}
\usepackage{cuted}
\usepackage{caption}
\usepackage{multirow}
\usepackage{xspace}
\usepackage{pifont}

\newcommand{\ours}{Molt Forcing\xspace}  %
\newcommand{\sink}{\mathcal{S}}
\newcommand{\win}{\mathcal{W}}
\newcommand{\cur}{\mathcal{C}}
\newcommand{\pin}{p_{\text{init}}}
\newcommand{\psw}{p_{\text{sw}}}
\newcommand{\pcold}{p_{\text{cold}}}
\newcommand{\ts}{t_{\text{sw}}}
\DeclareMathOperator{\erode}{erode}
\DeclareMathOperator{\fillholes}{fill}

%% file: authors.tex
\author{Hao Zhou$^{1}$ \quad Ziyang Xu$^{1}$ \quad Chenhan Jin$^{1}$ \quad Hao Qin$^{1}$ \quad Shengze Xu$^{1}$ \quad Tieyong Zeng$^{2}$\\[3pt]
{\normalsize $^{1}$The Chinese University of Hong Kong}\\
{\normalsize $^{2}$Beijing Normal-Hong Kong Baptist University}}

%% file: fig_en/teaser.tex
\begin{strip}
\centering
\includegraphics[width=\textwidth]{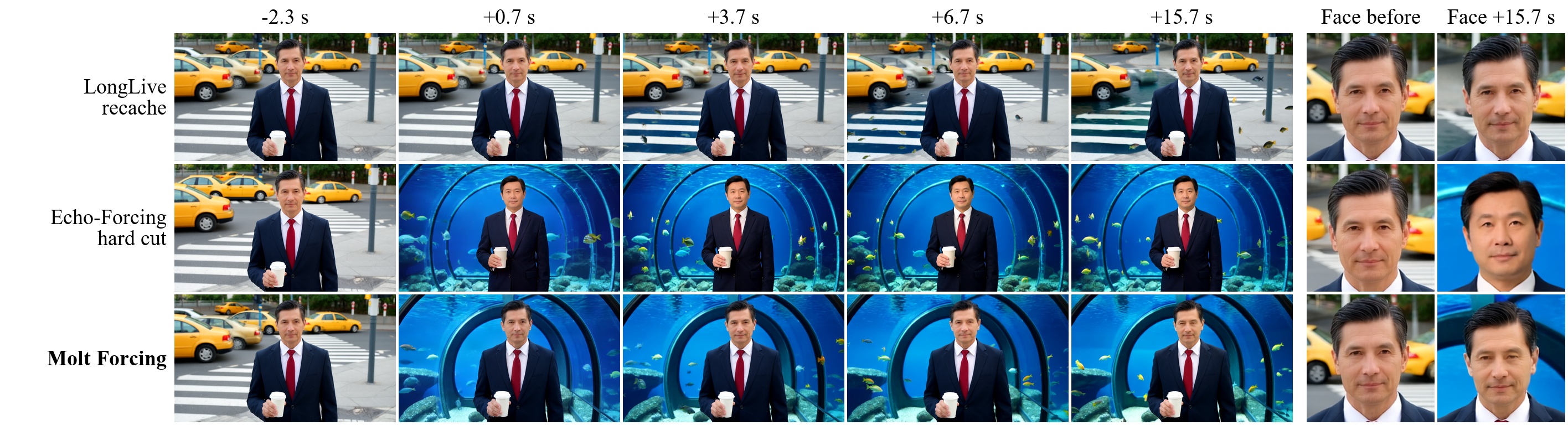}
\captionof{figure}{\textbf{The same history can preserve the subject and retain the old scene.}
The prompt switches from a city street to an aquarium tunnel (0\,s).
LongLive recache retains street objects alongside fish; Echo-Forcing hard cut changes the environment but changes the face; \ours shows the aquarium with closer resemblance to its pre-switch subject.
Right: manual face crops before the switch and at +15.7\,s. Echo uses its official cache configuration and therefore its own pre-switch video.
Businessman, seed 0, author-selected; rule-selected examples are in \cref{fig:qual}.}
\label{fig:teaser}
\end{strip}

%% file: sec_en/0_abstract.tex
\begin{abstract}
Switching scenes while preserving the original subject in streaming video generation exposes a conflict: cached history helps maintain the subject but also sustains old-scene objects that should disappear.
Through controlled interventions on text conditioning, denoising steps, the local window, and the attention sink, we identify how cached visual history keeps this obsolete content.
When the original frames remain in the cache, KV flush and sink recomputation still leave old-scene objects in the generated videos.
Replacing this history with target-scene history largely removes those objects, but also changes the subject, reducing face similarity to the pre-switch subject.
We then propose \ours, a training-free method that rebuilds scene history from a cold-start target scene while selectively retaining subject tokens from the original cache.
Across 16 scene families, \ours removes old-scene objects at least as reliably as an Echo-Forcing hard cut while keeping the subject much closer to its pre-switch appearance.
\end{abstract}

%% file: sec_en/1_intro.tex
\section{Introduction}
\label{sec:intro}
Autoregressive video diffusion models generate videos block by block, using cached visual history to maintain temporal consistency~\cite{yin2025causvid,huang2025selfforcing,yang2025longlive}.
In interactive generation, a scene change should preserve the original subject while updating its surroundings.
Moving a man holding a coffee cup from a street to an aquarium, for example, should preserve the man and his cup while removing taxis and street infrastructure.
Yet cached history contains both the subject and the old scene.
In \cref{fig:teaser}, recomputing cached states under the new prompt leaves the street in place, whereas an Echo-Forcing hard cut~\cite{echoforcing} changes the environment but also alters the subject's appearance.

Recent methods manage memory for scene transitions~\cite{echoforcing}, separate how subject and scene queries access history~\cite{tethermem}, or use reference-based anchors to preserve subject appearance~\cite{customforcing}.
We study which parts of the cache keep obsolete scene content after a prompt switch, and how to remove that content without losing information about the original subject.

We examine two cache components: an attention sink retaining initial frames and a local window retaining recent frames~\cite{xiao2024attentionsink,liu2026rollingforcing,yang2025longlive}.
Paired interventions hold the pre-switch video and generation noise fixed.
Changing the text condition leaves old objects in place, and so do two operations that keep native frames in the cache: flushing the window down to its last frame while keeping the sink~\cite{yesiltepe2026infinityrope}, and recomputing the sink under the new prompt.
Each clears 0 of 32 videos.
Shortening or recomputing history is not forgetting it.
Replacing the window and the sink together with target-scene history clears 31 of 32, but it also replaces the subject, and similarity to the pre-switch face drops.
We therefore edit cached history selectively, replacing scene content and keeping the subject.

\ours is a training-free method that handles subject and scene history separately.
At the switch, it introduces target-scene context from a separate cold-start generation and restricts reads from the original history to the subject region.
After the switch, the attention sink keeps native subject tokens, and the local window fills with newly generated frames.
In a follow-up intervention that holds cached content fixed, shifting the spatial rotary positional embeddings (RoPE) of history keys moves the generated background by the predicted amount, while the subject region moves less.

Subject continuity matters most at the switch, where viewers compare the subject directly across the cut.
A hard cut regenerates the subject at that point: 0.7\,s after the switch, Echo's face similarity to the pre-switch frame is .39, against .82 without a switch.
\ours reaches .70 at the same moment.
Across 16 scene families, it clears old objects in 30 of 32 videos (Echo: 28) and keeps a higher pre-switch face similarity than Echo through about 16\,s after the switch.

Our contributions are:
\begin{itemize}
\item \textbf{A controlled diagnosis} of how the local window and the attention sink sustain old-scene content, including spatial interventions showing that the generated background follows the position at which history is read.
\item \textbf{A comparison of history edits} showing that window flushing and sink recomputation clear no video, whereas replacing both components with target-scene history clears nearly all at a cost to subject resemblance.
\item \textbf{\ours}, a training-free method that selectively retains native subject history; it matches Echo-Forcing hard cut on old-object removal and preserves more of the pre-switch face.
\end{itemize}

%% file: sec_en/2_related.tex
\section{Related Work}
\label{sec:related}

\paragraph{Autoregressive video diffusion and sinks.}
CausVid~\cite{yin2025causvid} distills a bidirectional model into a causal few-step generator, while Self Forcing~\cite{huang2025selfforcing} trains on autoregressive rollouts conditioned on its own outputs.
LongLive~\cite{yang2025longlive} combines short-window attention with a frame-level sink and refreshes caches when prompts change.
Long-horizon stability is addressed through supervision of segments sampled from self-generated long videos~\cite{cui2026selfforcingpp}, rolling denoising with initial-frame attention sinks~\cite{liu2026rollingforcing,xiao2024attentionsink}, and sink expansion with temporal RoPE realignment and cache pruning~\cite{yi2026deepforcing}.
Persistent anchors can also constrain scene evolution: AdaState~\cite{dalva2026adastate} replaces the fixed first-frame anchor with an evolving latent state, while DySink~\cite{dysink} retrieves historical frames as dynamic sinks.

\paragraph{KV cache management and memory.}
Cache selection policies consider redundancy~\cite{densitykv}, anticipated query relevance~\cite{futureforcing}, newly appearing content~\cite{recapforcing}, and relevance with temporal diversity~\cite{tethercache}.
Memory mechanisms further differ in prompt-based retrieval~\cite{memflow}, object-centric organization~\cite{slotmemory}, and selective injection into memory-sensitive layers~\cite{ding2026layerrecall}.
TetherMem~\cite{tethermem} separates subject and scene queries and adjusts historical access by region and age, allowing scene progression while preserving subject recognizability.
For our task, scene progression must include removing specified old objects after a prompt switch, so we evaluate removal and subject preservation jointly.

\paragraph{Interactive prompt switching.}
Infinity-RoPE~\cite{yesiltepe2026infinityrope} retains the sink and last latent frame through KV Flush and uses RoPE Cut for shot transitions.
Other approaches warm-start cache recomputation from anchors~\cite{yang2026anchorforcing}, interpolate old- and new-prompt caches in a framework trained in two stages~\cite{chen2026groundedforcing}, or inject prompt semantics while adapting the local window~\cite{tan2026swift}.
Echo-Forcing~\cite{echoforcing} organizes anchors, compressed history, and recent frames into hierarchical memory with difference-aware decay.
It supports smooth transitions, hard cuts, and long-range scene recall; its hard-cut setting directly matches our task of changing the background while preserving the subject.
Among contemporaneous methods, SEGUE~\cite{rao2026segue} plans intermediate prompts from the current visual state and user instructions, and trains the generator to follow their schedule, while Weave Forcing~\cite{wang2026weaveforcing} selects subject and background references separately and masks historical KV tokens for compositional reuse.
Relative to these methods, we analyze how the sink and local window keep old-scene content and how replacing history affects the subject.
\ours then replaces the scene part of the cache with history from a cold-start target generation and keeps the native subject tokens.

\paragraph{Multi-shot generation and subject preservation.}
ShotStream~\cite{luo2026shotstream} and FilmWeaver~\cite{luo2026filmweaver} train generators with separate cross-shot and within-shot memories for visual consistency and temporal coherence.
External-reference methods refine inputs and sampling guidance~\cite{tpige} or combine image-based adaptation with value warping~\cite{vwarper}.
Contemporaneous Custom Forcing~\cite{customforcing} stores reference-based anchors in a persistent KV cache and amplifies their influence under identity drift.
These methods specify the desired subject through supplied images; our task starts with a subject already established in the generated video.
\ours instead takes its identity anchor from subject tokens in the model's own pre-switch attention sink, requiring no external reference images or additional training.

%% file: sec_en/3_prelim.tex
\section{Task and Preliminaries}
\label{sec:prelim}
\paragraph{Scene switching with subject preservation.}
Given a generated video prefix and a new scene prompt, the task is to continue the video in the target environment while retaining the established subject and its specified attributes.
Old objects designated for removal should disappear.
We study one known switch per video, without external subject-reference images.
Success requires both scene updating and subject preservation; retaining the old scene or regenerating a different subject satisfies only part of the task.

\paragraph{Block-wise autoregressive generation.}
Let $z_{1:N}$ denote a video's latent sequence.
The model generates one block $z_b=z_{[s,s+B)}$ at a time, where $s$ is its starting index and $B$ is the block length.
Each block undergoes $K$ denoising steps, with step $k$ predicting a clean latent $\hat{x}_0^{(k)}$.
After denoising, the model performs an additional forward pass at $t=0$ (the \emph{clean write}), storing the block's keys and values (K/V) in the cache for subsequent generation.

\paragraph{Read context.}
At each layer, self-attention for block $b$ reads the concatenated keys and values
\begin{equation}
  \mathrm{KV}_{b} = [\,\sink \,;\, \win_{b} \,;\, \cur_{b}\,],
  \label{eq:view}
\end{equation}
where $\sink$ contains K/V from the first $|\sink|$ latent frames (the attention sink~\cite{xiao2024attentionsink}), $\win_b$ from the most recent completed blocks (the local window), and $\cur_b$ from the current block.
In our LongLive configuration, these components span 3, 6, and 3 latent frames, respectively.
We call this concatenation the block's \emph{read context}; text conditioning enters separately through cross-attention.
Cache entries computed from the model's own pre-switch video are \emph{native}, and history generated from the target prompt alone in an empty cache is \emph{cold-start}.

\paragraph{Prompt switching.}
At the switch point $\ts$, the text condition changes from $\pin$ to $\psw$.
LongLive recomputes K/V for the latest 12 latent frames under the new prompt (KV recache), leaving the sink unchanged in its official interactive configuration~\cite{yang2025longlive}.
Echo-Forcing's hard cut discards the original sink and window, retaining only one frame of memory compressed from the previous scene~\cite{echoforcing}.
This old memory is cleared after one block, so subsequent sink and window contents come from frames generated after the switch.
Unless otherwise stated, we switch at latent frame $\ts=93$ (decoded frame 369, about 23\,s at 16\,fps), with fixed generation lengths before and after the switch.
Each latent frame spans four decoded frames (0.25\,s); we report all times in seconds relative to the switch, so +6.7\,s is the end of latent frame 119, and the pre-switch reference frame for identity lies 2.3\,s before the switch.

%% file: sec_en/4_diagnosis.tex
\section{Diagnosis: Updating Scenes while Retaining Subjects}
\label{sec:diagnosis}
\input{fig_en/diagnosis}

This section tests which cache components keep old-scene content after a switch, and what replacing them costs the subject; protocols are detailed in \cref{sec:setup} and supplement Secs.~B, F, and J.

\subsection{Replacing window and sink}
\label{sec:diag-window-sink}
We compare the local window and the sink in a $2\times2$ replacement experiment.
Each component either retains native history or reads K/V from a cold-start target video.
Within each pre-switch video, all four settings share the target prompt and generation noise and perform no recaching.
At 6.7\,s after the switch, the vision-language model (VLM) judge labels all three specified old objects absent in 0/32 videos with neither component replaced, 3/32 with only the sink replaced, 20/32 with only the window replaced, and 31/32 with both replaced (\cref{fig:diag}b; supplement Tab.~S9).
Window-only replacement varies across families: 3/12 videos in the six earlier families versus 17/20 in the ten newly added families.

Replacing only one component can also mix old and target scenes, as when fish appear over the old street.
Both cache components thus sustain old-scene content, and replacing them together clears it.
Across the 20 human videos, however, mean face similarity to the pre-switch reference frame at 6.7\,s falls from .734 with native history to .350 with both components replaced.
We therefore ask whether less extensive changes to history can remove the unwanted content.

\subsection{Edits that keep native history}
\label{sec:diag-text}
\label{sec:predictions}
Changing the text alone leaves the residue in place.
Rewriting negative old-object descriptions as target-environment descriptions and steering text embeddings leave the targeted objects in development probes; LongLive recaching with a prompt that names no old object clears none of 32 videos in the original blinded review (supplement Tab.~S1 and Sec.~G).
The same target prompts, used in cold-start probes without history, produce recognizable target environments with no old objects in all eight videos.
Wording still has an effect.
Objects named in negated sentences can reappear after an Echo hard cut, so our target prompts never name old objects.

We then test two operations that retain native visual content.
KV flush keeps the three-frame native sink and only the last native latent frame of the window.
Sink recomputation extends LongLive recaching to the sink, recomputing its K/V under the target prompt without changing its original latents.
Under the same three-object criterion at 6.7\,s, KV flush and sink recomputation each clear 0/32 videos, compared with 20/32 for window replacement (\cref{fig:diag}b).
Both operations leave native K/V in the read context, and both leave the old scene in place.

The old layout is already present at the first denoising step, but later steps can still change it.
In a development case (two seeds, four samples each), the old cabinet appears in the first-step prediction of every trial, yet reading target-reference K/V at steps 2--4 removes it (\cref{fig:diag}a).

\begin{figure}[t]
\centering
\includegraphics[width=0.93\linewidth]{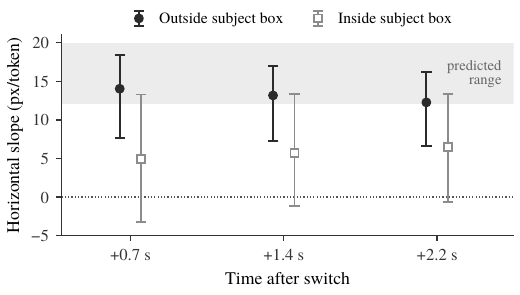}
\caption{\textbf{Spatial RoPE response.} Horizontal displacement slope for all-history shifts on eight fixed videos, with video-bootstrap 95\% CIs. The shaded band is the predicted 12--20 pixels/token range. The background outside the subject box follows the shift; the subject region moves less.}
\label{fig:spatial}
\end{figure}
\input{fig_en/method}

\subsection{Where history is read}
\label{sec:diag-copy}
To test spatial correspondence without replacing visual content, a follow-up experiment shifts the horizontal RoPE phase of pre-switch history keys on eight fixed seed-0 videos while holding the cached content fixed.
At 0.7/1.4/2.2\,s after the switch, the background outside the subject box moves by 14.0/13.2/12.3 pixels per token, within the predicted 12--20 range; inside the box, the displacement is smaller, at 4.9/5.7/6.5 (\cref{fig:spatial}).
Background placement thus follows the position at which history is read, while the subject region responds more weakly.

Regional replacement shows the same correspondence: moving a replacement boundary from left to right produces a vertical seam that moves with it (\cref{fig:diag}c; supplement Sec.~B).
A spatial mask therefore selects token positions, and old content stored at those positions is read together with the subject.
In development, retaining subject regions either kept old-background fragments along the subject contour or let identity drift (supplement Tab.~S2), which shaped the boundary and anchor choices in \cref{sec:method}.

%% file: fig_en/diagnosis.tex
\begin{figure*}[t]
\centering
\includegraphics[width=\textwidth]{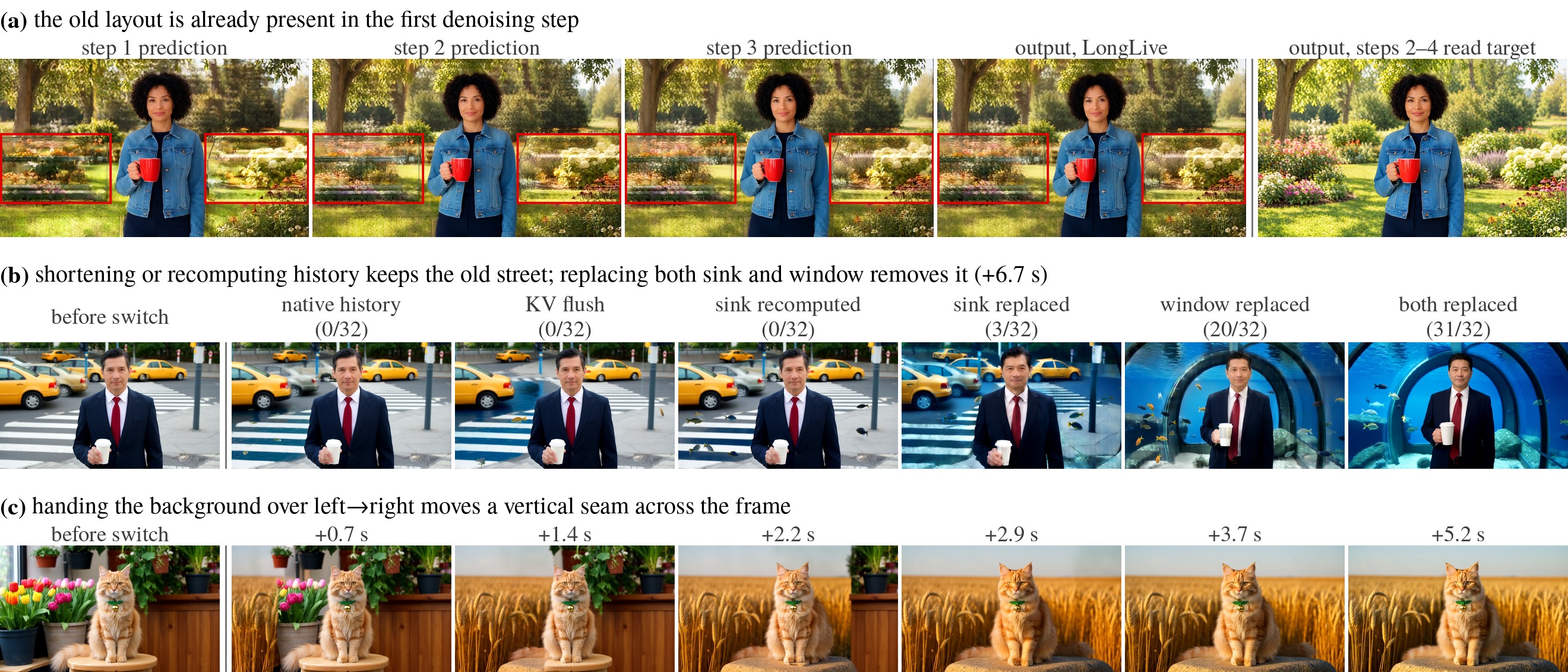}
\caption{\textbf{Diagnosis.}
(a) Development case (woman, caf\'e to garden), an inspected block after the switch. In the first denoising step's prediction, glass cabinets with the position and size of the old pastry display already appear on both sides (red boxes), although the block starts from noise. LongLive keeps them; reading the target scene at steps 2--4 removes them.
(b) One pre-switch video (businessman, seed 0) under six history operations, 6.7\,s after the switch. KV flush and sink recomputation keep native frames in the read context and keep the street; replacing only the sink adds fish over it; replacing the window or both reaches the aquarium. Brackets: videos out of 32 in which all three old objects are judged absent (\cref{tab:main}; supplement Tab.~S9).
(c) Cat example from the regional-replacement diagnostic. Handing the background to the target scene from left to right produces a vertical seam that moves with the handed-over region (\cref{sec:diag-copy}).}
\label{fig:diag}
\end{figure*}

%% file: fig_en/method.tex
\begin{figure*}[t]
\centering
\includegraphics[width=0.95\textwidth]{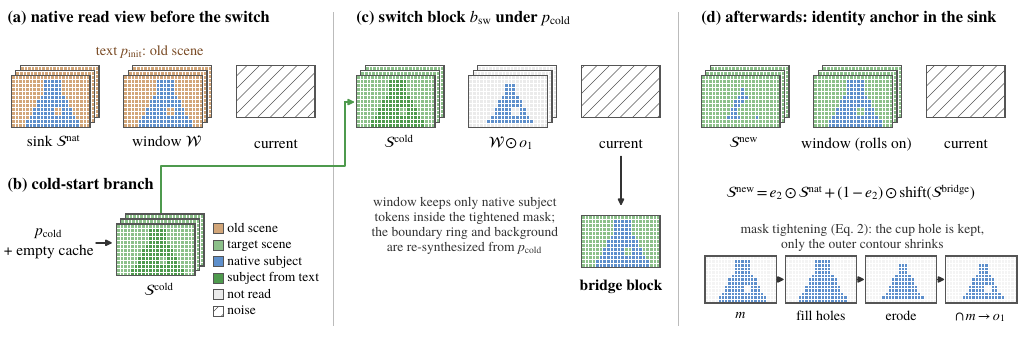}
\caption{\textbf{Overview of \ours} (schematic token grid).
(a) Before the switch, both the sink and the local window come from the old scene.
(b) At the switch, a cold-start branch generates the first $|\sink|$ frames of a video under $\pcold$ in an empty cache, giving a target-scene sink $\sink^{\text{cold}}$; its subject is generated from text and differs from the original one.
(c) The switch block uses $\sink^{\text{cold}}$ as the sink, and its window reads only native subject tokens inside the tightened outer contour ($o_1$, \cref{eq:outer-erode}); the boundary ring and the background are resynthesized under $\pcold$, giving the bridge block.
(d) Afterwards, the sink consists of the shifted bridge block plus the native-sink subject tokens in region $e_2$ (\cref{eq:anchor}), which act as the identity anchor; the window rolls as usual.
Bottom right: filling holes before eroding tightens only the outer contour, so the hole at a held object does not erode the subject outward.}
\label{fig:method}
\end{figure*}

%% file: sec_en/5_method.tex
\section{Method}
\label{sec:method}

\ours introduces target-scene history while retaining selected native subject tokens (\cref{fig:method}).
A cold-start branch supplies scene context, a restricted native window carries subject information through the switch, and a native subject sink provides a persistent reference as the window rolls.
\Cref{sec:ablation} evaluates these components.

\subsection{Cold-start target sink}
\label{sec:cold-sink}
At the switch, we generate the first $|\sink|$ latent frames of a video in a separate empty cache under a cold-start prompt $\pcold$, obtaining the cold-start K/V $\sink^{\text{cold}}$.
$\pcold$ is rewritten from $\psw$: it keeps the subject and its preserved attributes from $\pin$, drops transition wording such as ``gradually'', and names no old object (supplement Sec.~E).
The cold-start branch uses its own random state, leaving the main noise sequence untouched.
Unlike reusing a fixed target reference frame, this branch needs no external reference and samples scene content separately for each generated video.

\subsection{Boundary-tightened switch block}
\label{sec:bridge}
Rather than replacing the entire native window, we retain subject-region tokens during the switch to preserve access to the established subject.
At the switch, the main branch first performs the official LongLive recache under $\pcold$, which leaves the sink unchanged, and all routing below acts on the recached cache.
The switch block $b_{\text{sw}}$ starting at $\ts$ is generated with $\sink^{\text{cold}}$ as the sink and $\pcold$ as the text, and its local window reads only native tokens inside the subject region.
The subject mask $m$ comes from CLIPSeg~\cite{luddecke2022clipseg} on the subject phrase, refined by SAM~\cite{kirillov2023sam} and downsampled to the token grid.
Subject-region selection leaked old background at the contour in development (supplement Tab.~S2), so we tighten the outer contour:
\begin{equation}
  o_n = m \,\cap\, \erode\!\big(\fillholes(m),\, n\big),
  \label{eq:outer-erode}
\end{equation}
where $\fillholes$ fills interior holes (\eg, at the held red cup) and $\erode(\cdot,n)$ applies $n$ erosion steps, so only the outer contour shrinks.
Intersecting with $m$ leaves the holes unread, so objects inside them are redrawn from $\pcold$.
The switch block reads $\win^{\text{sw}} = \win_{b_{\text{sw}}} \odot o_1$, and the one native block still in the next block's window is also read only inside $o_1$.
The boundary ring is thus resynthesized under $\pcold$, and the subject interior keeps native content; we call the result the \emph{bridge} block.

\begin{table*}[t]
\centering\small
\setlength{\tabcolsep}{5pt}
\caption{\textbf{Eight-condition comparison.} Residue: fraction of present labels over nine times and three old objects per video. Cleared: videos with all three objects absent at 6.7\,s. Face: ArcFace similarity at 6.7\,s to the condition's own pre-switch frame. Brackets: pointwise 95\% family-bootstrap CIs. The two column groups have different denominators.}
\label{tab:main}
\begin{tabular}{lcccc}
\toprule
 & \multicolumn{2}{c}{All 32 videos / 16 families} & \multicolumn{2}{c}{20 human videos / 10 families} \\
\cmidrule(lr){2-3}\cmidrule(lr){4-5}
Method & Residue $\downarrow$ & Cleared $\uparrow$ & Residue $\downarrow$ & Face $\uparrow$ \\
\midrule
Native history & .949 [.902, .988] & 0/32 & .981 [.948, 1.000] & .734 [.647, .811] \\
Replace sink & .714 [.571, .844] & 3/32 & .774 [.626, .904] & .402 [.350, .452] \\
Replace window & .087 [.043, .135] & 20/32 & .081 [.030, .135] & .363 [.320, .402] \\
Cold restart & .013 [.000, .036] & 31/32 & .000 [.000, .000] & .350 [.308, .391] \\
Echo hard cut~\cite{echoforcing} & .043 [.000, .105] & 28/32 & .002 [.000, .006] & .373 [.313, .428] \\
KV flush & .929 [.861, .983] & 0/32 & .980 [.946, 1.000] & .730 [.638, .806] \\
Sink recomputation & .938 [.881, .985] & 0/32 & .980 [.946, .998] & .725 [.635, .804] \\
\ours & .024 [.000, .059] & 30/32 & .000 [.000, .000] & .587 [.509, .667] \\
\bottomrule
\end{tabular}
\end{table*}

\subsection{Native subject identity anchor}
\label{sec:anchor}
Window retention alone provides only temporary access to the native subject tokens, since those frames eventually leave the window.
We therefore retain a subject reference in the sink.
After the bridge block, the sink becomes the K/V of the bridge block (RoPE positions shifted from $\ts$ to $0$), except that the subject positions are replaced by the corresponding tokens of the native sink $\sink^{\text{nat}}$:
\begin{equation}
  \sink^{\text{new}} = e_2 \odot \sink^{\text{nat}} + (1-e_2) \odot \mathrm{shift}\big(\sink^{\text{bridge}}\big),
  \label{eq:anchor}
\end{equation}
where $e_2=\erode(m^{\text{nat}},2)$ is the subject mask on the native sink frames after 2 erosion steps; all layers use this sink.
Once the native pre-switch frames ($\ts+6$) and the bridge block ($\ts+9$) leave the window, these sink tokens are the only native subject K/V left in the read context.
Other anchor sources let identity drift in development: subject tokens from recent pre-switch frames or the bridge block, and native subject tokens read in only some layers.

\paragraph{Cost.}
\ours adds one cold-start block per switch (4 denoising calls and 1 clean write, about 1.1\,s per call); afterwards every block reads the same length as the original model.
The subject mask is computed once offline on pre-switch frames.

%% file: sec_en/6_experiments.tex
\section{Experiments}
\label{sec:experiments}

We evaluate whether selective history use combines old-object removal with improved resemblance to the pre-switch subject.
We first compare the joint outcomes, then examine the components, temporal range, and motion.
The unified table also reports the diagnostic operations from \cref{sec:diagnosis} under the same protocol.
\subsection{Setup}
\label{sec:setup}
\paragraph{Model and data.}
We use LongLive~\cite{yang2025longlive} with Wan2.1-T2V-1.3B weights~\cite{wan2025}, four denoising steps, $832\times480$ output, and 16\,fps.
The switch is at latent $\ts=93$ (about 23\,s into the video).
A scene family is one subject with an initial and a target scene; each family is generated with two seeds, one video each.
The method was developed on two families (rabbit and woman) and frozen before validation on six more.
The main set contains these six and ten additional families first generated for this comparison: 16 families, 32 videos, including 20 human and 12 non-human videos.
Each family specifies three old objects and subject attributes to preserve.
We keep all 32 videos, including seven whose pre-switch video misses a requested attribute; the subset without them is analyzed in supplement Sec.~G.
Target prompts are manually written, shared across methods, and never name old objects (supplement Sec.~E).

\paragraph{Comparisons.}
The main comparison has eight conditions: native history without recache, sink-only replacement, window-only replacement, cold restart, \ours, Echo-Forcing hard cut, KV flush, and sink recomputation.
KV flush retains the native sink and the last native latent frame; sink recomputation recomputes the native sink under the target text as part of recache.
LongLive's own recache and Echo's smooth mode are in the supplementary comparison.
All LongLive-based conditions share the same pre-switch videos and the same CPU-generated noise, indexed by absolute block and step.
Echo uses LongLive weights with its official window/sink configuration (21/1 versus 12/3) and therefore generates its own pre-switch videos.
Face similarity is measured against each condition's own pre-switch video.

\paragraph{Evaluation.}
At nine fixed times from 0.7 to 6.7\,s after the switch, spaced 0.75\,s apart, Qwen2.5-VL-7B~\cite{bai2025qwen25vl} labels each of the three old objects present, absent, or unclear.
The residual rate is the fraction labeled present among these 27 questions; unclear remains in the denominator (zero in this batch).
A video is cleared when all three objects are labeled absent at 6.7\,s.
Against 1,392 earlier labels from a blinded LLM-based viewer that inspected the same frames, leave-one-family calibration gives balanced accuracy .956 and $\kappa=.917$ (supplement Sec.~F).
Face similarity is ArcFace~\cite{deng2019arcface} cosine similarity after YuNet~\cite{wu2023yunet} detection, against a frame 2.3\,s before the switch.
We first average the two seeds within each family, then weight families equally.
Paired effects are computed within videos before family aggregation; brackets give pointwise 95\% percentile bootstrap intervals over 2,000 family resamples.
Predictions for the follow-up experiments were frozen before generation.
CLIP text matching, DINOv2~\cite{oquab2024dinov2} background similarity, and the six VBench~\cite{huang2024vbench} dimensions are reported in supplement Secs.~F and G.

\subsection{Joint removal and preservation}
\label{sec:main-results}
\begin{figure}[t]
\centering
\includegraphics[width=\linewidth]{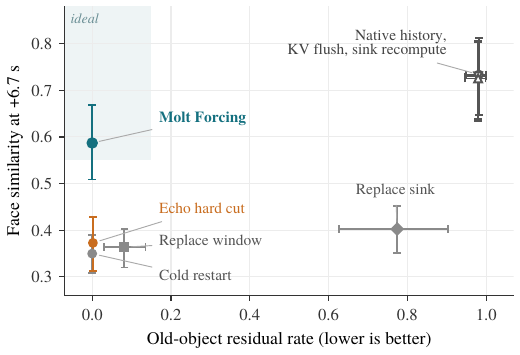}
\caption{\textbf{Residue and face similarity on the same 20 human videos.} Bars are pointwise family-bootstrap 95\% intervals on each axis. The task asks for the upper left: low residue and high similarity. Native history keeps the face but also the old scene; reset methods remove the scene at lower similarity.}
\label{fig:joint}
\end{figure}

\input{fig_en/qualitative_col}
\cref{tab:main} reports all 32 videos and the 20 human videos separately; \cref{fig:joint} uses the human videos on both axes, and \cref{fig:qual} shows two examples.
Among the diagnostic controls, retaining native content leaves high residue, and full replacement lowers face similarity.
\ours combines low residue with higher face similarity than the reset methods.

On the human videos, the paired face-similarity advantage of \ours over Echo at 6.7\,s is .214 (\cref{sec:identity-time}).
On removal and text matching the two methods are on par: the paired human residual-rate difference is $-.002$ [$-.006$, .000], and CLIP text--frame similarity over all videos at 6.7\,s is .378 for \ours and .382 for Echo, with paired difference $-.004$ [$-.008$, .002].
Paired forced-choice judgments on the same frames agree with the face-similarity ordering: the blinded LLM-based viewer preferred \ours over the hard cut in 84, tied in 4, and preferred the hard cut in 12 of 100 pairs (89/6/5 against cold restart), and an author check preferred it in 37 of 40 pairs with 3 ties against each reset method (supplement Secs.~F, G).

\subsection{Components}
\label{sec:ablation}
\cref{tab:components} adds the components of \ours one at a time to a cold restart.
Reading native subject tokens through the window during the switch raises face similarity from .350 to .528, and keeping native subject tokens in the sink raises it to .617 but brings back collage-like patches of old background.
Eroding the anchor and tightening the window contour remove these patches while keeping most of the gain (.587).

\begin{table}[t]
\centering\small
\setlength{\tabcolsep}{2.5pt}
\caption{\textbf{Components} (32 videos at 6.7\,s; face similarity on the 20 human videos). Clear: cleared videos. Collage and leak: videos with the artifact (unclear in brackets). Each row adds to the one above; the last row is \ours. Full table with LongLive recache in supplement Tab.~S8.}
\label{tab:components}
\begin{tabular}{lcccc}
\toprule
Configuration & Clear$\uparrow$ & Collage$\downarrow$ & Leak$\downarrow$ & Face$\uparrow$ \\
\midrule
Cold restart & 31 & 0 (0) & 0 (1) & .350 \\
+ native subject window & 29 & 1 (1) & 1 (4) & .528 \\
+ native subject sink & 29 & 3 (5) & 0 (7) & .617 \\
+ eroded anchor $e_2$ & 29 & 1 (2) & 1 (5) & .596 \\
+ tight window $o_1$ (ours) & 30 & 0 (0) & 0 (1) & .587 \\
\bottomrule
\end{tabular}
\end{table}

\subsection{Identity over time}
\label{sec:identity-time}
\begin{figure}[t]
\centering
\includegraphics[width=0.90\linewidth]{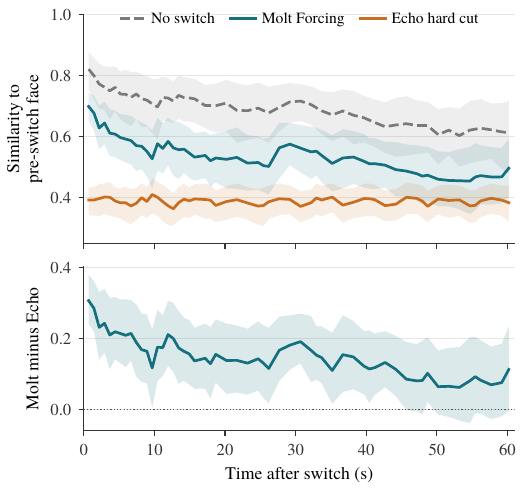}
\caption{\textbf{Face similarity through 60\,s after the switch.} Top: similarity to each condition's own pre-switch frame. Bottom: paired \ours--Echo difference. Shading gives pointwise family-bootstrap 95\% intervals over 10 families.}
\label{fig:curve}
\end{figure}

A long-horizon extension follows all 20 human videos under \ours, Echo hard cut, and a no-switch reference to 60\,s after the switch (\cref{fig:curve}).
Measured against the pre-switch frame, the paired advantage over Echo is .214 [.137, .307] at 6.7\,s, .137 [.059, .230] at 15.7\,s, and .113 [$-.004$, .234] at 60\,s; the interval stays above zero through about 16\,s after the switch.
At 60\,s, similarities are .496 for \ours, .384 for Echo, and .613 without a switch, and \ours leads in 8 of 10 families.
Measured instead against each output's own frame at 0.7\,s, the \ours--Echo difference is $-.204$ [$-.305$, $-.118$]: \ours stays closer to the original subject, while Echo stays closer to the subject it regenerates at the cut.
Echo also redraws unprompted clothing details such as sleeve length, which \ours keeps (supplement Sec.~J).
Neither method has an old object labeled present at 60\,s (0/20).

\subsection{Motion}
\label{sec:motion-limits}
Motion is evaluated on 15 action pairs that the backbone completes from a cold start, each with two seeds.
VLM completion counts for cold restart, \ours, \ours without its persistent anchor, Echo, and no switch are 12/14/14/15/20 out of 30.
The gap lies between switching and not switching rather than between methods with and without the native anchor.

%% file: fig_en/qualitative_col.tex
\begin{figure}[t]
\centering
\includegraphics[width=\linewidth]{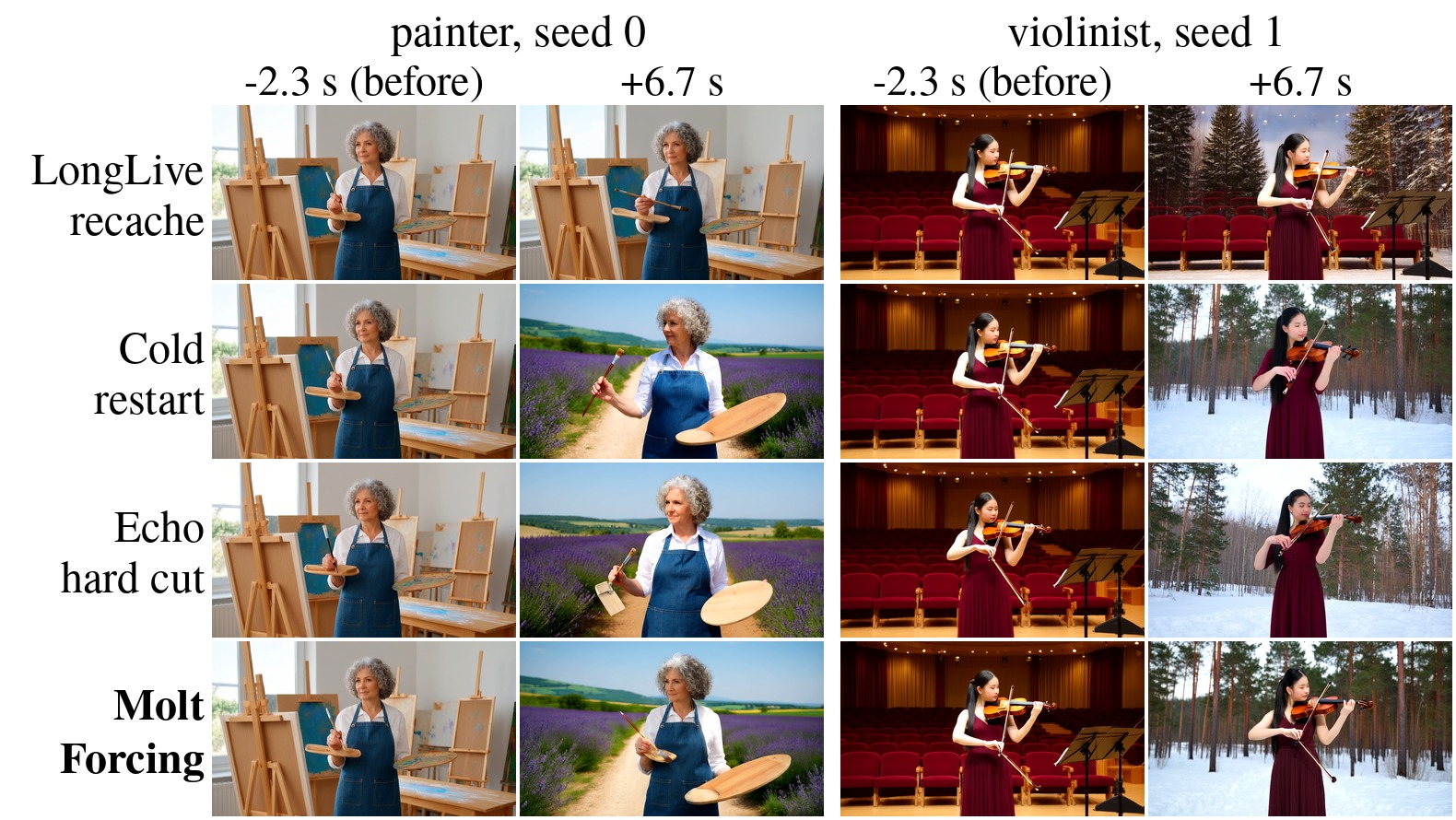}
\caption{\textbf{Examples chosen by a fixed rule}: the two human videos whose \ours--Echo face-similarity gap over 2.2--6.7\,s after the switch is closest to the median. LongLive recache keeps the old studio and concert hall (in the violinist, snowy trees appear above the seats). The other three reach the new scene; \ours keeps the pre-switch subject appearance, whereas the cold restart and Echo regenerate it from text. More frames in supplement Sec.~A.}
\label{fig:qual}
\end{figure}

%% file: sec_en/7_conclusion.tex
\section{Limitations and Conclusion}
\label{sec:conclusion}
\paragraph{Limitations.}
\emph{Scope.} The full evaluation uses one backbone, one subject, and one known switch per video, with manually written target prompts and offline masks whose time is not included in the reported cost.
\ours produces cuts; gradual transitions, multiple switches, returns to earlier scenes, and reliable automatic triggering remain untested or unresolved.
Motion coverage is limited: the large-motion stratum contains only two canine action pairs and no humans, so the motion results are not an equivalence test.

\emph{Mechanism.} The diagnosis does not establish a unique origin of old-scene content, nor that replacement is the only way to remove it.
Inside the subject box, all three RoPE displacement intervals include zero, and the predicted 75\% window-only share of displacement is not robustly established (supplement Sec.~J).
Spatial masks select positions rather than pure identity features, so leakage and subject damage remain possible; objects inside mask holes are constrained only by $\pcold$, not reproduced exactly.
Mask and anchor choices were selected in development, and the final component step changes both the window mask and recache, so it is not a single-factor ablation.

\emph{Evidence.} The six earlier validation families were later reused for diagnosis, and the unified table reuses the diagnostic operations, so these are not independent replications.
Echo has a different cache configuration and its own pre-switch videos, limiting causal attribution across the two systems.
VLM labels at sampled frames do not certify continuous absence, and ArcFace is a proxy rather than human identity ground truth: author review found abnormal face shape in both chef seeds despite similar identity.
Automatic labels, LLM-based viewer judgments, and non-blind author checks are distinct sources, none an independent human study.
Intervals are pointwise, and aggregation choices were frozen after generation but before each aggregation.
The identity advantage at 60\,s is not established: its interval includes zero, failing the preregistered positive-lower-bound condition (family sign test $p=.109$); why the gap narrows is unresolved, and the clothing observations are unblinded and unmeasured.

\paragraph{Conclusion.}
The same cached history carries both the subject and the old scene.
Our interventions show that both the local window and the sink keep old-scene content. Flushing and recomputing history leave that content in place, and joint replacement removes it at a cost to facial similarity.
\ours, which requires no training, introduces target-scene history and selectively retains native subject tokens.
It achieves low old-object residue and higher pre-switch-reference facial similarity than Echo hard cut; the paired interval for this advantage stays above zero through about 16\,s after the switch.
Editing cached history selectively, rather than keeping or discarding it whole, removes the old scene after a prompt switch while keeping more of the original subject.

%% file: sec/8_llm_statement.tex
\section*{LLM Usage Statement}
We used LLM assistants (Claude, Anthropic; Codex, OpenAI) for coding, literature organization, discussion of experiment design, drafting, editing, and translation.
Automatic old-object labels use Qwen2.5-VL-7B~\cite{bai2025qwen25vl}; action labels also use a VLM and are reported as proxies.
Separate LLM-based viewer sessions produced the original blinded old-object and identity judgments and diagnostic static checks.
An author manually judged a fixed subset without LLM assistance; subsequent author video review was non-blind and is reported separately.
These sources are not presented as an independent human preference study.
The authors are responsible for verifying the reported results and conclusions.

%% file: supplement.tex
\setcounter{section}{0}
\setcounter{figure}{0}
\setcounter{table}{0}
\renewcommand{\theHsection}{supp.\arabic{section}}
\renewcommand{\theHtable}{supp.\arabic{table}}
\renewcommand{\theHfigure}{supp.\arabic{figure}}
\renewcommand{\thesection}{\Alph{section}}
\renewcommand{\thetable}{S\arabic{table}}
\renewcommand{\thefigure}{S\arabic{figure}}
\newsavebox{\suppbox}
\newcommand{\suppfit}[1]{\sbox{\suppbox}{#1}\ifdim\wd\suppbox>\linewidth\resizebox{\linewidth}{!}{\usebox{\suppbox}}\else\usebox{\suppbox}\fi}

\raggedbottom
\maketitlesupplementary

\section{Qualitative examples}
\label{sec:supp-video}
\begin{figure*}[t]
\centering\includegraphics[width=\textwidth]{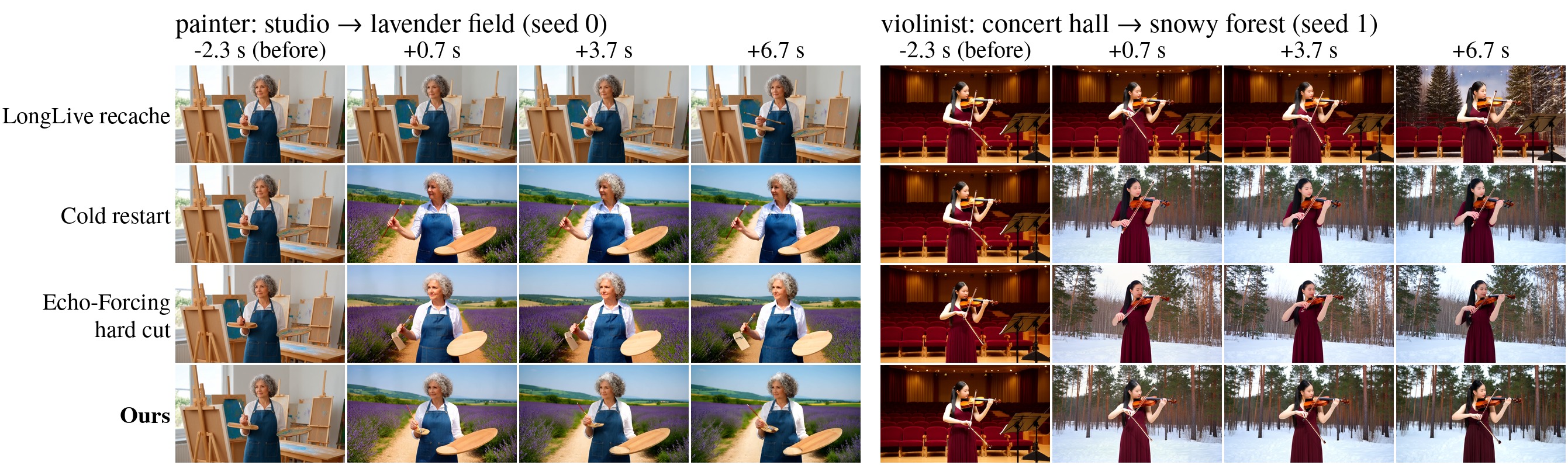}
\caption{\textbf{Examples selected by a fixed metric rule.} Among 20 human videos, these two have the Molt--Echo face-similarity gap over 2.2--6.7\,s after the switch closest to the median: painter seed 0 and violinist seed 1. The LongLive recache row retains old-scene structure; clearing methods change the environment with different subject appearances.}
\label{fig:supp-qual}\end{figure*}

Echo-Forcing~\cite{echoforcing} uses its official cache configuration and therefore has its own pre-switch video.
Main-paper Figure 1 shows the author-selected street-to-aquarium family, seed 0, with the original fixed time columns and manual face crops; rule-selected examples are in \cref{fig:supp-qual}.

\section{Diagnostic chain}
\label{sec:supp-chain}
Development uses two families (rabbit and woman), two seeds each, and multiple continuation samples.
Repeated samples from one pre-switch video count as one task.
Tables~\ref{tab:supp-text} and \ref{tab:supp-chain} summarize the text tests and the sequence of interventions.
Each development experiment used its own recorded pass criteria, which cover part of the full scene-change task.
Early experiments used an oracle target-reference block; later ones start from the model's own pre-switch video.

\begin{table*}[t]
\centering\small
\caption{\textbf{Text interventions.} History is retained in the first three rows, absent in the fourth, and cleared before the last.}
\label{tab:supp-text}
\begin{tabular}{@{}p{6.6cm}p{9.6cm}@{}}
\toprule
Intervention & Observation \\
\midrule
Replace negative old-object wording with an end-state description & Old furniture remains in all four rabbit continuations. \\
Positive, negative, and random text-embedding steering & The woman's display remains under all directions in both seeds. \\
Target prompt without old-object names; LongLive recache or Echo smooth & Clearance at 5.2 and 6.7\,s in the original blinded review is 0/32 for either method. \\
Generate a target scene without history & Recognizable target environment in 8/8; no old object in inspected frames. \\
Delete a negated sentence naming old objects after Echo hard cut & Final background progress increases from .514 to .891; naming a negated object can still evoke it. \\
\bottomrule\end{tabular}
\end{table*}

\begin{table*}[t]
\centering\small
\caption{\textbf{Development diagnostics}, in the order they were run. The first three rows use an oracle target-reference state; the fourth compares it with the model's own pre-switch video, which all later rows use.}
\label{tab:supp-chain}
\begin{tabular}{@{}p{6.2cm}p{10.0cm}@{}}
\toprule
Intervention & Result \\
\midrule
Temporal position and phase of a target-reference bank & Initial effects are sensitive to relabeling; sustained phase alignment passes only 1/4. Temporal alignment alone is insufficient. \\
Decode denoising predictions; replace history at selected steps & Structure appears at step 1 (8/8 in the woman probe); interventions at steps 2--4 can still remove it. \\
Repeat late-step routing, add stay controls, then withdraw it & 3/4 switch videos pass; both woman seeds regain the cabinet after withdrawal. \\
Separate clean writes, sink, and text; compare oracle/native state & The same cold sink and text work differently under the oracle state and the model's own pre-switch video. The starting state matters. \\
Replace window and sink; retain a subject box; omit sink handoff & Full replacement clears but changes identity; box retention can splice scenery; omitting sink handoff brings ghosts back. \\
Subject-only switch window and bridge sink & Old objects are removed, but identity drifts later; subject positions also carry non-identity content. \\
Native-sink versus recent-frame anchors; layer subsets & Native sink improves identity but leaks background; no tested layer subset suffices. \\
Anchor erosion and outer-boundary tightening & Boundary changes reduce leakage; $o_1$ window plus $e_2$ anchor is frozen as Molt Forcing. \\
\bottomrule\end{tabular}
\end{table*}

We leave out an earlier action-reversion experiment, because returning to a prior action is a different outcome from retaining an old scene object.
\paragraph{Composition transfer and regional replacement.}
With both the window and sink replaced by cold-start content, development probes give a mean absolute background difference from the cold-start preview of 0.03--0.09, versus 0.29--0.39 for native generation.
Using a fixed target reference also makes backgrounds from different samples converge toward its composition.
In a later regional-replacement study, 12 pre-switch videos with two left-to-right schedules give 24 outputs with moving vertical seams.
Two scattered-region schedules on the same videos leave old structures visible in the inspected late frames of all 24 outputs.
The four vertical bands first reach 50\% background progress in strict left-to-right order in 11/12 and 10/12 outputs for the two sweep schedules, respectively; one latter-schedule chef output has three bands that do not reach the threshold.
Neither schedule produces a clean gradual transition.
The fixed-content, inside/outside-box RoPE measurements are reported separately in Sec.~J.

\section{Implementation}
\label{sec:supp-impl}
\paragraph{Read views.}
Latent indices below count from the start of the video; the switch is at 93, and each latent frame spans 0.25\,s.
LongLive generates three latent frames per block, with $30\times52=1560$ tokens per frame.
At block start $s$, attention reads the sink (0:3), two previous blocks ($s-6:s-3$ and $s-3:s$), and the current block ($s:s+3$).
There are four denoising calls at 1000, 937.5, 833.3, and 625, followed by a clean write at 0.
A read hook supplies modified K/V views at all 30 layers and all five calls; the routing hook does not mutate the underlying cache. Official recache and normal clean writes still update it.

\paragraph{Block schedule.}
At $s=93$, the sink reads cold-start K/V, and the two native window blocks (87:93) retain only $o_1$ tokens; excluded tokens are removed from attention, not zeroed.
At $s=96$, the sink reads the combined bridge/native-subject sink; the remaining native block 90:93 is still restricted to $o_1$, and the bridge block is read normally.
At $s\ge99$, only the sink view is changed. Native pre-switch frames leave the window at 99 and the bridge at 102.
After the bridge clean write, its keys are temporally shifted from position 93 to 0 using RoPE; values are unchanged.
Positions inside $e_2$ are then replaced in a copy by native sink K/V.
The original sink remains available in the cache.

\paragraph{Recache and cold start.}
Molt first performs official LongLive recache on 81:93 under the target prompt, which leaves the native sink unchanged.
Cold restart does not recache, so the last component step changes both recache and the window mask.
The cold branch independently generates one block at position 0, with four denoising calls and one clean write; Cold restart and Molt use the same cold-start K/V.
Its separate random state does not consume main-branch noise.

\paragraph{Noise.}
In the original comparison, bf16 noise is generated on CPU and saved before generator calls; input SHA checks and actual re-noising sequences are recorded.
The post-switch initial-noise seed is 5904 (sample index 4).
For family index $c$, re-noising uses $760000+100c+\mathrm{seed}$, cold initial noise uses $66000+1000c+100\mathrm{seed}+\mathrm{sample}$, and cold re-noising uses $770000+100c+\mathrm{seed}$.
Methods index noise by absolute block and step; the first 27 post-switch re-noising tensors match within each video.

\begin{figure*}[t]
\centering\begin{minipage}{.88\linewidth}\footnotesize
\begin{verbatim}
e2, o1 = erode(native_sink_mask, 2), outer_erode(native_window_mask, 1)
cold = generate_block(empty_cache, p_cold, start=0, noise=cold_noise)
recache(cache, p_cold, frames=81:93)              # native sink unchanged
for s in 93, 96, ..., end:
    for t in 1000, 937.5, 833.3, 625, 0:         # t=0 performs clean write
        if s == 93: read_hook(sink=cold, window=native[87:93] restricted to o1)
        elif s == 96: read_hook(sink=S_new, slot1=native[90:93] restricted to o1)
        else: read_hook(sink=S_new)
        x = generator(x, t, p_cold, cache)
    if s == 93:
        B = cache.kv[93:96].copy(); B.k = rope_shift(B.k, 93 -> 0)
        S_new = B.copy(); S_new[e2] = cache.kv[0:3][e2]
\end{verbatim}
\end{minipage}
\caption{\textbf{One-switch inference schedule.} Spatial routing changes the read view; ordinary recache and clean-write operations retain their usual roles.}
\label{fig:supp-code}\end{figure*}

\section{Subject masks}
\label{sec:supp-mask}
Masks are computed offline on 33 pre-switch frames: the first nine frames of the video and the last 24 frames before the switch.
CLIPSeg uses the family's subject phrase and a threshold at half its maximum probability.
The 1st/99th percentiles of the selected pixel coordinates define a box for SAM ViT-B.
Among three SAM candidates, we maximize IoU with the CLIPSeg region plus 0.5 times SAM's predicted IoU.
Masks from decoded frames $4L-3$ through $4L$ are unioned for latent frame $L$ (latent frame 0 uses only the first frame).
Tokens with at least .25 coverage on the $16\times16$ pixel grid are selected, followed by one $3\times3$ dilation.
The sink anchor $e_2$ uses two four-neighbor erosions.
The window mask $o_1=m\cap\mathrm{erode}(\mathrm{fill}(m),1)$ tightens only the external boundary; intersecting with $m$ leaves internal holes unread.
An initial region below 50 pixels or an empty required mask counts as a method failure and is neither repaired nor excluded; none occurs in the 32 videos.
Original token counts are: $e_2$ on the last sink frame, median 226 (56--353); $o_1$ on the last pre-switch latent frame, median 297 (64--454).

\begin{figure*}[t]
\centering\includegraphics[width=.9\linewidth]{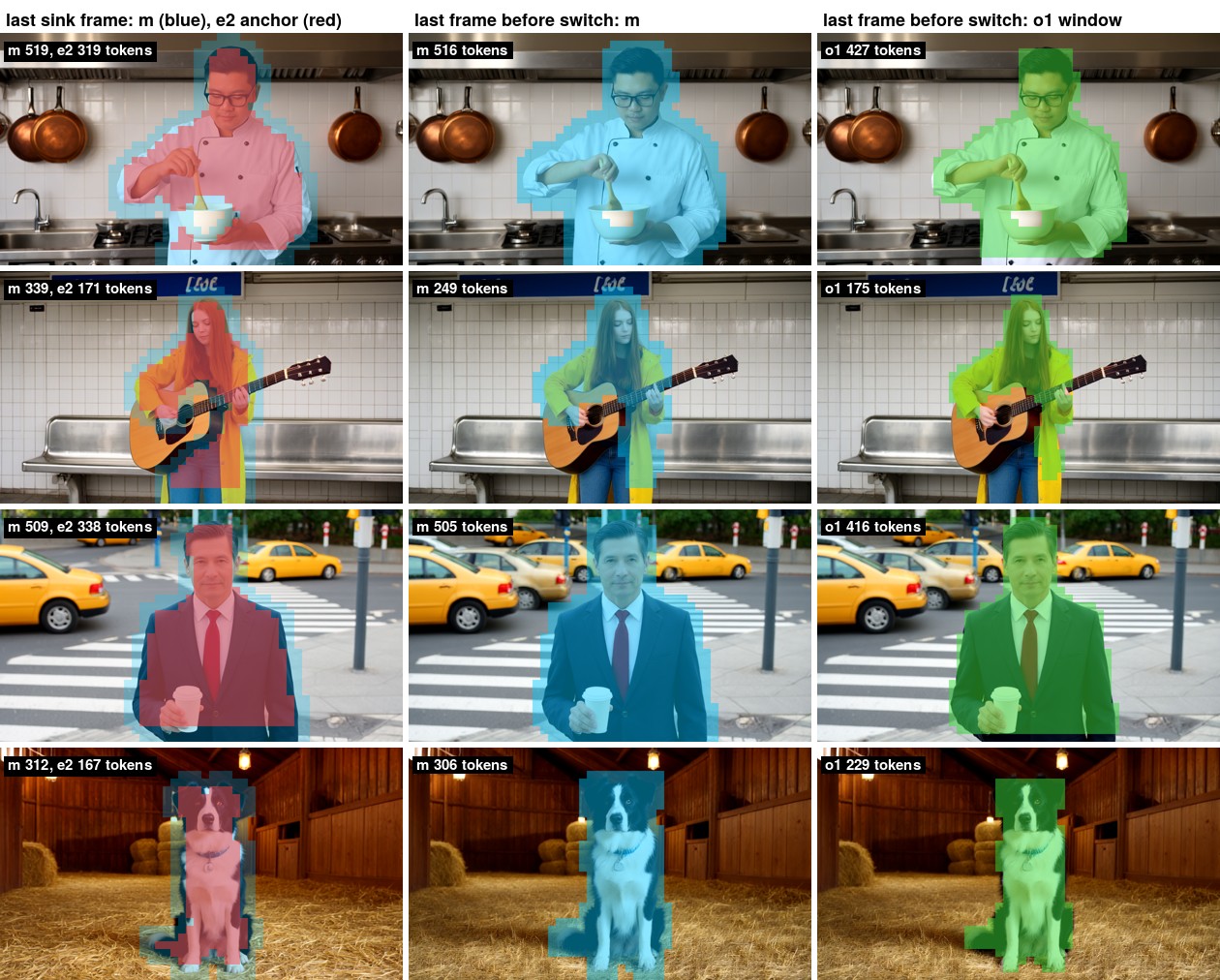}
\caption{\textbf{Masks for chef, guitarist, businessman, and collie, seed 0.} Left: native sink mask $m$ (blue) and eroded anchor $e_2$ (red) on the last sink frame. Middle: $m$ on the last frame before the switch. Right: $o_1$ on that frame. Labels give token counts. The guitarist anchor excludes the sides of the long hair.}
\label{fig:supp-masks}\end{figure*}

\section{Prompts and scene families}
\label{sec:supp-prompts}
The initial prompt names the subject, preserved attributes, source scene, and three old objects.
The manually written target prompt retains the subject/attribute description, substitutes the target environment, omits gradual-transition wording, and never names old objects.
Prompts exclude Echo parser characters \texttt{; [ ] \# @}.
Both seeds use identical text; videos whose pre-switch part misses an attribute are kept as generated.
The ten new families were first generated after these prompts were fixed.
\input{supp_prompts_en}

\section{Evaluation protocols and evidence sources}
\label{sec:supp-protocol}
\paragraph{Blinded LLM-based viewer.}
The viewer is a separate LLM session, uninvolved in method development, reading anonymized images, questions, and object/attribute names without method names or automatic metrics.
Qualification used 64 pre-switch panels, one per video and system, at 2.3 and 0.1\,s before the switch. At least two old objects and the subject with kept attributes had to be visible in the later frame; unknown did not qualify.
Semantic judging used 160 continuations from five switching methods, with frames at 5.2 and 6.7\,s after the switch as primary and 0.7\,s plus a nine-point strip as context.
Transparent but recognizable ghosts counted as present; out-of-view or ambiguous content remained unknown.
Identity used 200 paired questions (20 human videos, two opponents, five times), 12 no-switch positive controls, and 12 swapped retests, with full images and $320\times320$ face crops.
Labels and image-display records were frozen before unblinding.
One disconnected qualification attempt had no saved labels and was repeated; one interrupted semantic session resumed from saved labels.

\paragraph{Author subset.}
One author manually judged a fixed 40-video semantic subset (240 labels) and 80 identity questions, with additional controls and retests, without LLM assistance.
The author did not assess qualification, kept attributes, damage, or collage.
Semantic agreement was 238/240 ($\kappa=.98$); three-way identity agreement was 66/80.
The author gave Molt win/tie/loss 37/3/0 against each reset opponent.

\paragraph{Post-unblinding correction.}
For man seed 0, the viewer marked a right-edge wooden structure unknown for both cold restart and Molt.
The author identified it as a target-scene bench post connected to the backrest; both conditions were changed to absent, and collage to no, in a separate adjudication record.
The original viewer files stay frozen, and the correction applies only to the original viewer table.
\begin{figure}[t]
\centering\includegraphics[width=\linewidth]{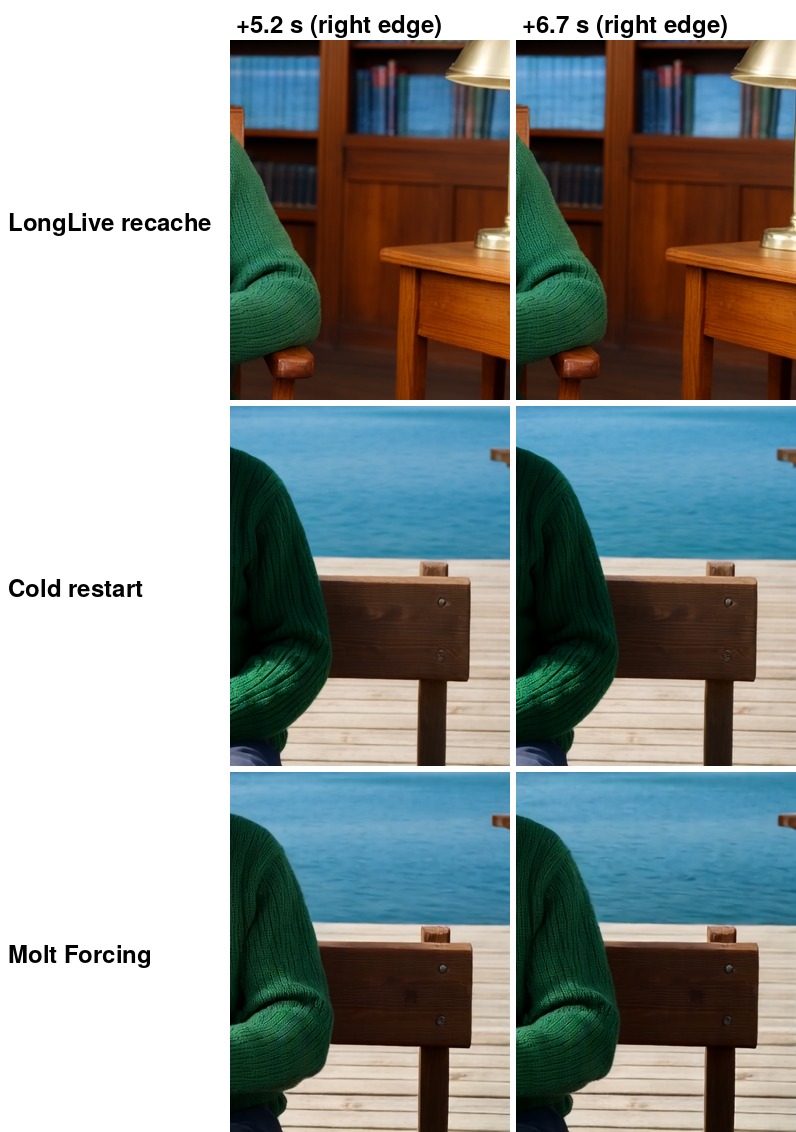}
\caption{\textbf{Adjudication crop, man seed 0, right edge ($x\ge552$), 5.2 and 6.7\,s after the switch.} Recache retains the reading table; the reset and Molt outputs show the target bench post.}
\label{fig:supp-adjudication}\end{figure}

\paragraph{Automatic object labels.}
Qwen2.5-VL-7B-Instruct answers present/absent/unclear with fixed prompts and greedy decoding, using only visible content.
The original leave-one-family calibration against 1,392 frozen labels from the blinded LLM viewer gives balanced accuracy .956, $\kappa=.917$, residual recall .935, and endpoint clearance agreement .938.
The weakest rabbit fold has balanced accuracy .771 and $\kappa=.54$; the weakest held-out residual recall is .79 for dog.
The new main comparison uses nine times and three objects, 27 questions per video.
Unclear labels stay in the denominator as non-absent; none occurs in this batch.
The pooled source has 8,832 logical labels, with exact-input reuse recorded separately; long-run images never replace short-run images of the same condition.

\paragraph{Statistics.}
Two seeds are averaged within families, then families receive equal weight.
Paired differences are formed before aggregation; percentile CIs use 2,000 resamples and random seed 880701.
Family sign tests, rather than seed-level tests, are primary.
All intervals are pointwise, without multiplicity correction.
Missing data stay missing, with explicit denominators.
The prediction document preceded the video runs; metric aggregation, cached-feature aggregation, VLM aggregation, and the missing-CLIP completion each have their own implementation record, frozen later.

\clearpage
\onecolumn
\paragraph{Two background definitions.}
Let $s,t$ be cosine similarity of background DINO features to the source and cold reference.
The original progress is $\max(0,1-s)/(\max(0,1-s)+\max(0,1-t))$, retaining its $10^{-4}$ degeneracy rule.
The added progress is $(1-s)/(1-s_{\mathrm{source,cold}})$ without clipping.
Background features exclude the union of source and cold subject boxes.
The registered window-replacement value of about .82 uses only the original definition.
CLIP uses the same target prompt across the eight main conditions, full RGB frames, and a 77-token limit, with token counts retained. The no-switch reference is scored against its initial prompt.

\section{Additional quantitative tables}
\label{sec:supp-tables}
The original blinded-viewer comparison and its eligible subset below keep their original labels and two-time clearance definition, separate from the nine-time VLM rates of the main paper.
The seven ineligible pre-switch videos are student seeds 0/1 (notebook not open), producer 0/1 (headphones on ears rather than neck), parrot 0/1 (yellow wing tips unclear), and violinist 0 (hair not in the required low ponytail).

\input{supp_tables_en}

The viewer's Molt win/tie/loss over all 20 human videos is 84/4/12 against Echo and 89/6/5 against cold restart.
In the eligible subset it is 60/4/11 and 64/6/5, respectively.
Both positive controls and swapped retests are 12/12 in the original viewer protocol.
Disagreement concentrates in guitarist, hiker seed 0, and producer (\cref{sec:supp-failures}).
VBench covers the five original switching methods and two windows (\cref{tab:supp-vbench}).

\section{Failure cases}
\label{sec:supp-failures}
The guitarist seed-0 output shifts red hair toward brown; the anchor misses the hair's lateral regions. We suspect the autumn background color but did not test it.
Hiker seed 0 changes hat and face shape despite favorable ArcFace; the viewer chose the opponent on every original question for that seed, whereas the author never did and used two ties.
Collie seed 0 has two tail-like structures in the Molt output, recorded as damage.
In one development example using Self-Forcing weights but the LongLive cache configuration, late white fragments appear under Molt; skipping the subject anchor in the last block removes them in a replay of that state.

The later author video review reports abnormal face shape in both chef seeds at displayed continuation times 00:52 and 00:44, respectively, while still labeling identity as the same.
High face similarity or a same-identity label therefore does not rule out facial damage.

\begin{center}\begin{minipage}{.82\linewidth}
\centering\includegraphics[width=\linewidth]{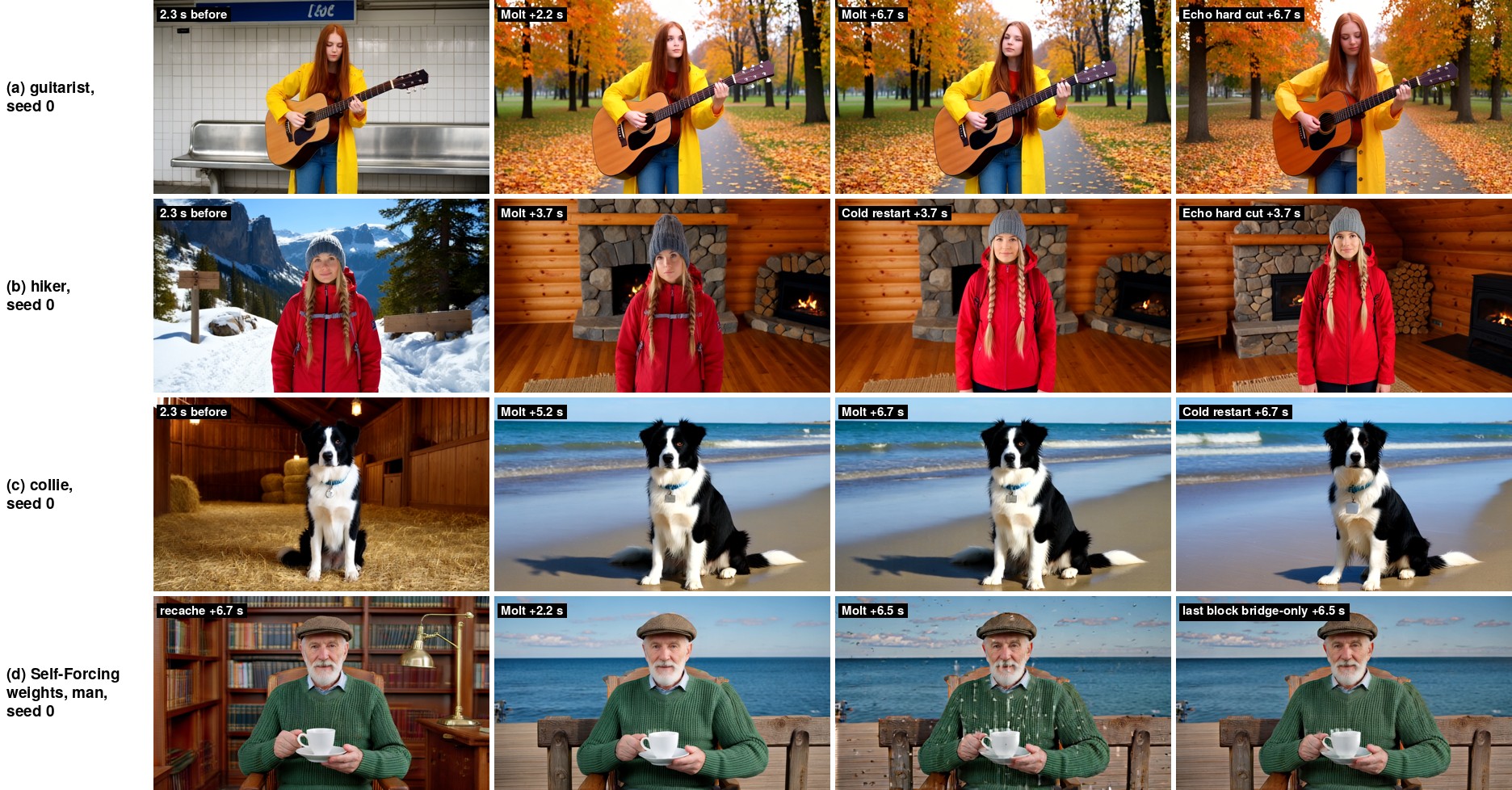}
\captionof{figure}{\textbf{Original failure examples.} (a) Guitarist: hair-color change. (b) Hiker: hat/face-shape change. (c) Collie: double tail. (d) Self-Forcing-weight development example: late fragments, removed by the tested final-block read change. The first tile shows the pre-switch frame or a baseline; Echo's pre-switch videos differ. The later chef observations are reported separately in text.}
\label{fig:supp-failures}\end{minipage}\end{center}

\section{Computational cost}
\label{sec:supp-cost}
Molt adds one cold-start block (four denoising calls plus one clean write) relative to LongLive recache, and six calls relative to Echo hard cut including recache.
Subsequent blocks retain the original read length.
\input{supp_cost_en}
Counts use the original short-run lengths.
Echo's original runner also regenerates its pre-switch video (155 calls); that execution overhead is excluded from the post-switch table.
The original 32-video comparison records 26,074 generator calls and 472 VAE decodes, totaling about 29,309 program seconds, including loading and decoding.
Jobs used one 32\,GB RTX 5090 GPU, as recorded for the shared environment, and passed a 30\,GiB peak-memory assertion.
Mask computation is included in jobs and was not separately timed.
Later evaluation explicitly reuses compatible cached features and labels; new model calls and engineering checks are separately recorded in the evidence manifest.

\section{Follow-up predictions, long horizons, and motion}
\label{sec:supp-followup}
\input{supp_followup_en}

%% file: supp_prompts_en.tex
\begin{table*}[t]
\centering\scriptsize
\setlength{\tabcolsep}{3pt}
\caption{\textbf{16 scene families}. Earlier: six families first used for validation after method freezing; new: ten first generated for the main comparison. Identity: A = face similarity and forced choice (human); D = descriptive DINO only (non-human). Prompts and attributes are unchanged.}
\label{tab:supp-families}
\begin{tabular}{@{}llp{2.3cm}p{5.2cm}p{4.4cm}c@{}}
\toprule
Family & Group & Source $\to$ Target & Kept attributes & Old objects & Identity \\
\midrule
dog\_petshop\_meadow & earlier & petshop $\to$ meadow & golden retriever; red neck bandana; one blue rubber ball near the front paws; plausible sitting posture & shelf of pet-food bags; colorful checkout counter; pet-shop walls & D \\
robot\_workshop\_gallery & earlier & workshop $\to$ gallery & small yellow robot; rectangular blue visor; one red circular chest emblem; two arms and two legs; plausible standing support & tool-covered pegboard; red tool cabinet; workshop walls & D \\
man\_library\_pier & earlier & library $\to$ pier & elderly man with a short white beard; brown flat cap; green knit sweater; one white porcelain teacup on a saucer held in both hands; plausible seated support & tall wooden bookshelves; brass reading lamp; wooden reading table & A \\
guitarist\_subway\_park & earlier & subway platform $\to$ autumn park & young woman with long straight red hair; yellow raincoat; blue jeans; one acoustic guitar; right hand strumming, left hand on the neck; plausible standing posture & white tiled wall; blue station name sign; metal platform benches & A \\
chef\_kitchen\_rooftop & earlier & kitchen $\to$ rooftop & man with short black hair and black-rimmed glasses; white chef jacket; one white ceramic bowl held in the left arm; wooden spoon in the right hand stirring; plausible standing posture & stainless steel range hood; hanging copper pots; white tiled kitchen wall & A \\
cat\_florist\_wheatfield & earlier & flower shop $\to$ wheat field & fluffy orange tabby cat; green collar; one small golden bell on the collar; plausible upright sitting posture & buckets of tulips; wooden shop counter; hanging potted plants & D \\
violinist\_concert\_snowforest & new & concert hall $\to$ snowy forest & young woman with long straight black hair in a low ponytail; long burgundy dress; one wooden violin under the chin; bow in the right hand; plausible standing posture & rows of red velvet theater seats; golden stage curtains; black music stands & A \\
painter\_studio\_lavender & new & art studio $\to$ lavender field & middle-aged woman with short curly gray hair; denim apron over a white shirt; one paintbrush in the right hand; wooden palette in the left hand; plausible standing posture & wooden easels with canvases; paint-splattered wooden table; tall studio window & A \\
student\_lecturehall\_lakeshore & new & lecture hall $\to$ mountain lake shore & young man with short brown hair; round black glasses; gray hoodie; one open red notebook held in both hands; plausible seated support & large green chalkboard; rows of wooden lecture desks; white projector screen & A \\
businessman\_street\_aquarium & new & city street $\to$ aquarium tunnel & man in his forties with neatly combed black hair; navy suit with white shirt and red tie; one white paper coffee cup in the right hand; plausible standing posture & yellow taxis; white crosswalk stripes; traffic light pole & A \\
hiker\_mountain\_cabin & new & snowy mountain trail $\to$ log cabin & woman with a long blond braid; red waterproof jacket; gray knit beanie; red backpack on her shoulders; plausible standing posture & snow-covered pine trees; rocky mountain peaks; wooden trail sign & A \\
producer\_studio\_canyon & new & recording studio $\to$ desert canyon & man with a shaved head and a full dark beard; orange hoodie; black headphones around the neck; plausible seated support & large mixing console; gray acoustic foam walls; microphone on a stand & A \\
reader\_train\_sunflower & new & train carriage $\to$ sunflower field & elderly woman with silver hair in a neat bun; round gold-rimmed glasses; red cardigan; one open hardcover book held in both hands; plausible seated support & train windows; metal luggage rack; blue patterned train seats & A \\
collie\_barn\_beach & new & barn $\to$ sunny beach & black-and-white border collie; blue collar; one silver name tag on the collar; plausible sitting posture & stacked hay bales; wooden horse stalls; hanging oil lanterns & D \\
parrot\_jungle\_livingroom & new & tropical jungle $\to$ modern living room & bright green parrot; red forehead; yellow wing tips; plausible perched posture on a wooden perch & large palm leaves; small waterfall; hanging green vines & D \\
panda\_zoo\_snowbamboo & new & zoo enclosure $\to$ snowy bamboo forest & giant panda; one green bamboo stalk held in the front paws; plausible sitting posture & gray concrete rock wall; wooden climbing frame; metal fence & D \\
\bottomrule
\end{tabular}
\end{table*}

\paragraph{Complete prompts.} Each family lists the CLIPSeg phrase, initial prompt $\pin$ and target prompt $\pcold$.

{\scriptsize
\noindent\textbf{dog\_petshop\_meadow} (CLIPSeg: ``a golden retriever'')\par
\noindent$\pin$: A continuous realistic medium-wide shot of a golden retriever wearing a red bandana around its neck, sitting on the floor inside a brightly lit pet shop. A tall shelf of pet-food bags and a colorful checkout counter are clearly visible behind the dog against the shop walls. One blue rubber ball rests on the floor between its front paws. Its ears, face, bandana, front paws, ball and sitting posture are clearly visible. The dog remains calmly sitting with natural breathing and subtle head and ear movements. The camera framing and indoor lighting remain steady.\par
\noindent$\pcold$: A continuous realistic medium-wide shot of a golden retriever wearing a red bandana around its neck, sitting in an open grassy meadow with wildflowers, rolling green hills and trees in the distance. One blue rubber ball rests on the grass between its front paws. Its ears, face, bandana, front paws, ball and plausible sitting posture are clearly visible. It remains calmly sitting with natural breathing and subtle head and ear movements. Natural daylight and steady camera framing.\par\medskip

\noindent\textbf{robot\_workshop\_gallery} (CLIPSeg: ``a small yellow robot'')\par
\noindent$\pin$: A continuous realistic medium-wide shot of a small yellow robot with a rectangular blue visor and one red circular emblem on its chest, standing on a wooden workbench inside a tidy workshop. A tool-covered pegboard and a red tool cabinet are clearly visible behind the robot against the workshop walls. Its complete head, visor, chest emblem, two arms, two hands, two legs, feet and standing support are clearly visible. It remains standing with subtle small head and arm movements. The camera framing and indoor lighting remain steady.\par
\noindent$\pcold$: A continuous realistic medium-wide shot of a small yellow robot with a rectangular blue visor and one red circular emblem on its chest, standing with plausible support on a simple white display pedestal in a spacious modern art gallery with plain white walls, polished pale flooring and framed abstract pictures behind it. Its complete head, visor, chest emblem, two arms, two hands, two legs and feet are clearly visible. It remains standing with subtle small head and arm movements. Soft gallery lighting and steady camera framing.\par\medskip

\noindent\textbf{man\_library\_pier} (CLIPSeg: ``an elderly man'')\par
\noindent$\pin$: A continuous realistic medium-wide shot of an elderly man with a short white beard wearing a brown flat cap and a green knit sweater, seated on a wooden chair inside a quiet library. Tall wooden bookshelves filled with books, a brass reading lamp and a wooden reading table are clearly visible behind him. He holds one white porcelain teacup on a saucer at chest height with both hands. His face, beard, cap, sweater, both hands, teacup, saucer and seated posture are clearly visible. He remains calmly seated with natural breathing and subtle head movements. The camera framing and warm lighting remain steady.\par
\noindent$\pcold$: A continuous realistic medium-wide shot of an elderly man with a short white beard wearing a brown flat cap and a green knit sweater, seated with plausible support on a simple wooden bench on a sunny wooden seaside pier with calm blue sea and clear sky behind him. He holds one white porcelain teacup on a saucer at chest height with both hands. His face, beard, cap, sweater, both hands, teacup, saucer and seated posture are clearly visible. He remains calmly seated with natural breathing and subtle head movements. Natural daylight and steady camera framing.\par\medskip

\noindent\textbf{guitarist\_subway\_park} (CLIPSeg: ``a young woman'')\par
\noindent$\pin$: A continuous realistic medium-wide shot of a young woman with long straight red hair wearing a yellow raincoat and blue jeans, standing on a subway platform and playing one acoustic guitar. A white tiled wall, a blue station name sign and metal platform benches are clearly visible behind her. Her right hand strums the strings while her left hand holds the guitar neck. Her face, hair, raincoat, both hands, guitar and standing posture are clearly visible. She keeps playing with natural strumming motion and subtle head movements. The camera framing and cool station lighting remain steady.\par
\noindent$\pcold$: A continuous realistic medium-wide shot of a young woman with long straight red hair wearing a yellow raincoat and blue jeans, standing in an autumn park with orange and red trees and a path covered with fallen leaves behind her, playing one acoustic guitar. Her right hand strums the strings while her left hand holds the guitar neck. Her face, hair, raincoat, both hands, guitar and plausible standing posture are clearly visible. She keeps playing with natural strumming motion and subtle head movements. Soft daylight and steady camera framing.\par\medskip

\noindent\textbf{chef\_kitchen\_rooftop} (CLIPSeg: ``a chef'')\par
\noindent$\pin$: A continuous realistic medium-wide shot of a chef with short black hair and black-rimmed glasses wearing a white chef jacket, standing inside a busy restaurant kitchen. A stainless steel range hood, hanging copper pots and a white tiled kitchen wall are clearly visible behind him. He holds one white ceramic bowl against his chest with his left arm and slowly stirs it with a wooden spoon in his right hand. His face, glasses, jacket, both hands, bowl, spoon and standing posture are clearly visible. He keeps stirring with natural hand motion and subtle head movements. The camera framing and bright kitchen lighting remain steady.\par
\noindent$\pcold$: A continuous realistic medium-wide shot of a chef with short black hair and black-rimmed glasses wearing a white chef jacket, standing on an open rooftop terrace at sunset with a city skyline and an orange sky behind him. He holds one white ceramic bowl against his chest with his left arm and slowly stirs it with a wooden spoon in his right hand. His face, glasses, jacket, both hands, bowl, spoon and plausible standing posture are clearly visible. He keeps stirring with natural hand motion and subtle head movements. Warm sunset light and steady camera framing.\par\medskip

\noindent\textbf{cat\_florist\_wheatfield} (CLIPSeg: ``an orange tabby cat'')\par
\noindent$\pin$: A continuous realistic medium-wide shot of a fluffy orange tabby cat wearing a green collar with one small golden bell, sitting upright on a low wooden stool inside a flower shop. Buckets of colorful tulips, a wooden shop counter and hanging potted plants are clearly visible behind the cat. Its ears, face, striped fur, collar, bell, front paws and sitting posture are clearly visible. The cat remains calmly sitting with natural breathing and subtle head and ear movements. The camera framing and soft shop lighting remain steady.\par
\noindent$\pcold$: A continuous realistic medium-wide shot of a fluffy orange tabby cat wearing a green collar with one small golden bell, sitting upright with plausible support on a flat stone in a golden wheat field under a clear evening sky. Its ears, face, striped fur, collar, bell, front paws and sitting posture are clearly visible. It remains calmly sitting with natural breathing and subtle head and ear movements. Warm evening light and steady camera framing.\par\medskip

\noindent\textbf{violinist\_concert\_snowforest} (CLIPSeg: ``a young woman'')\par
\noindent$\pin$: A continuous realistic medium-wide shot of a young woman with long straight black hair tied in a low ponytail, wearing a long burgundy dress, standing on the stage of a concert hall and playing one wooden violin held under her chin with a bow in her right hand. Rows of red velvet theater seats, golden stage curtains and black music stands are clearly visible around her. Her face, hair, dress, both hands, violin, bow and standing posture are clearly visible. She keeps playing slowly with gentle bow movements and subtle head movements. The camera framing and warm stage lighting remain steady.\par
\noindent$\pcold$: A continuous realistic medium-wide shot of a young woman with long straight black hair tied in a low ponytail, wearing a long burgundy dress, standing with plausible support on snow-covered ground in a quiet snowy pine forest clearing under a pale winter sky and playing one wooden violin held under her chin with a bow in her right hand. Her face, hair, dress, both hands, violin, bow and standing posture are clearly visible. She keeps playing slowly with gentle bow movements and subtle head movements. Soft winter daylight and steady camera framing.\par\medskip

\noindent\textbf{painter\_studio\_lavender} (CLIPSeg: ``a woman'')\par
\noindent$\pin$: A continuous realistic medium-wide shot of a middle-aged woman with short curly gray hair wearing a denim apron over a white shirt, standing inside an art studio and holding one paintbrush in her right hand and a wooden palette in her left hand. Wooden easels with canvases, a paint-splattered wooden table and a tall studio window are clearly visible behind her. Her face, hair, apron, both hands, paintbrush, palette and standing posture are clearly visible. She remains calmly standing with natural breathing and subtle head movements. The camera framing and soft daylight from the window remain steady.\par
\noindent$\pcold$: A continuous realistic medium-wide shot of a middle-aged woman with short curly gray hair wearing a denim apron over a white shirt, standing with plausible support on a dirt path in a blooming purple lavender field with distant hills under a clear blue sky, holding one paintbrush in her right hand and a wooden palette in her left hand. Her face, hair, apron, both hands, paintbrush, palette and standing posture are clearly visible. She remains calmly standing with natural breathing and subtle head movements. Bright natural daylight and steady camera framing.\par\medskip

\noindent\textbf{student\_lecturehall\_lakeshore} (CLIPSeg: ``a young man'')\par
\noindent$\pin$: A continuous realistic medium-wide shot of a young man with short brown hair and round black glasses wearing a gray hoodie, seated on a wooden chair inside a university lecture hall and holding one open red notebook with both hands. A large green chalkboard, rows of wooden lecture desks and a white projector screen are clearly visible behind him. His face, glasses, hair, hoodie, both hands, notebook and seated posture are clearly visible. He remains calmly seated with natural breathing and subtle head movements. The camera framing and even indoor lighting remain steady.\par
\noindent$\pcold$: A continuous realistic medium-wide shot of a young man with short brown hair and round black glasses wearing a gray hoodie, seated with plausible support on a large flat rock on a calm mountain lake shore with pine trees and snowy peaks in the distance, holding one open red notebook with both hands. His face, glasses, hair, hoodie, both hands, notebook and seated posture are clearly visible. He remains calmly seated with natural breathing and subtle head movements. Natural daylight and steady camera framing.\par\medskip

\noindent\textbf{businessman\_street\_aquarium} (CLIPSeg: ``a man in a suit'')\par
\noindent$\pin$: A continuous realistic medium-wide shot of a man in his forties with neatly combed black hair wearing a navy suit, a white shirt and a red tie, standing on a city sidewalk and holding one white paper coffee cup in his right hand. Yellow taxis, white crosswalk stripes and a traffic light pole are clearly visible behind him. His face, hair, suit, tie, both hands, coffee cup and standing posture are clearly visible. He remains calmly standing with natural breathing and subtle head movements. The camera framing and bright daylight remain steady.\par
\noindent$\pcold$: A continuous realistic medium-wide shot of a man in his forties with neatly combed black hair wearing a navy suit, a white shirt and a red tie, standing with plausible support inside an aquarium tunnel with curved glass walls, deep blue water and swimming fish around him, holding one white paper coffee cup in his right hand. His face, hair, suit, tie, both hands, coffee cup and standing posture are clearly visible. He remains calmly standing with natural breathing and subtle head movements. Soft blue aquarium light and steady camera framing.\par\medskip

\noindent\textbf{hiker\_mountain\_cabin} (CLIPSeg: ``a woman hiker'')\par
\noindent$\pin$: A continuous realistic medium-wide shot of a woman with a long blond braid wearing a red waterproof jacket and a gray knit beanie, with a red backpack on her shoulders, standing on a snowy mountain trail. Snow-covered pine trees, rocky mountain peaks and a wooden trail sign are clearly visible behind her. Her face, braid, beanie, jacket, backpack straps, both hands and standing posture are clearly visible. She remains calmly standing with natural breathing and subtle head movements. The camera framing and bright winter daylight remain steady.\par
\noindent$\pcold$: A continuous realistic medium-wide shot of a woman with a long blond braid wearing a red waterproof jacket and a gray knit beanie, with a red backpack on her shoulders, standing with plausible support on the wooden floor of a cozy log cabin interior with a stone fireplace, wooden log walls and a warm woven rug. Her face, braid, beanie, jacket, backpack straps, both hands and standing posture are clearly visible. She remains calmly standing with natural breathing and subtle head movements. Warm firelight and steady camera framing.\par\medskip

\noindent\textbf{producer\_studio\_canyon} (CLIPSeg: ``a bearded man'')\par
\noindent$\pin$: A continuous realistic medium-wide shot of a man with a shaved head and a full dark beard wearing an orange hoodie, with black headphones around his neck, seated on a black stool inside a recording studio. A large mixing console, gray acoustic foam walls and a microphone on a stand are clearly visible behind him. His face, beard, head, hoodie, headphones, both hands resting on his knees and seated posture are clearly visible. He remains calmly seated with natural breathing and subtle head movements. The camera framing and dim warm studio lighting remain steady.\par
\noindent$\pcold$: A continuous realistic medium-wide shot of a man with a shaved head and a full dark beard wearing an orange hoodie, with black headphones around his neck, seated with plausible support on a low red rock in a desert canyon with red sandstone cliffs under a bright clear sky. His face, beard, head, hoodie, headphones, both hands resting on his knees and seated posture are clearly visible. He remains calmly seated with natural breathing and subtle head movements. Bright desert sunlight and steady camera framing.\par\medskip

\noindent\textbf{reader\_train\_sunflower} (CLIPSeg: ``an elderly woman'')\par
\noindent$\pin$: A continuous realistic medium-wide shot of an elderly woman with silver hair in a neat bun, wearing round gold-rimmed glasses and a red cardigan, seated inside a passenger train carriage and holding one open hardcover book with both hands. Train windows, a metal luggage rack and blue patterned train seats are clearly visible around her. Her face, glasses, hair, cardigan, both hands, book and seated posture are clearly visible. She remains calmly seated with natural breathing and subtle head movements. The camera framing and soft daylight through the windows remain steady.\par
\noindent$\pcold$: A continuous realistic medium-wide shot of an elderly woman with silver hair in a neat bun, wearing round gold-rimmed glasses and a red cardigan, seated with plausible support on a white wooden bench in a sunflower field under a clear summer sky, holding one open hardcover book with both hands. Her face, glasses, hair, cardigan, both hands, book and seated posture are clearly visible. She remains calmly seated with natural breathing and subtle head movements. Bright summer daylight and steady camera framing.\par\medskip

\noindent\textbf{collie\_barn\_beach} (CLIPSeg: ``a border collie'')\par
\noindent$\pin$: A continuous realistic medium-wide shot of a black-and-white border collie wearing a blue collar with one silver name tag, sitting upright on a straw-covered floor inside a wooden barn. Stacked hay bales, wooden horse stalls and hanging oil lanterns are clearly visible behind the dog. Its ears, face, black-and-white fur, collar, name tag, front paws and sitting posture are clearly visible. The dog remains calmly sitting with natural breathing and subtle head and ear movements. The camera framing and warm barn lighting remain steady.\par
\noindent$\pcold$: A continuous realistic medium-wide shot of a black-and-white border collie wearing a blue collar with one silver name tag, sitting upright with plausible support on wet sand on a sunny sandy beach with gentle waves and a clear blue sky. Its ears, face, black-and-white fur, collar, name tag, front paws and sitting posture are clearly visible. It remains calmly sitting with natural breathing and subtle head and ear movements. Bright natural daylight and steady camera framing.\par\medskip

\noindent\textbf{parrot\_jungle\_livingroom} (CLIPSeg: ``a green parrot'')\par
\noindent$\pin$: A continuous realistic medium-wide shot of a bright green parrot with a red forehead and yellow wing tips, perched on a wooden branch perch in a tropical jungle. Large palm leaves, a small waterfall and hanging green vines are clearly visible behind the parrot. Its head, beak, eyes, red forehead, green feathers, yellow wing tips, feet and perched posture are clearly visible. The parrot remains calmly perched with natural breathing and subtle head movements. The camera framing and soft green-filtered daylight remain steady.\par
\noindent$\pcold$: A continuous realistic medium-wide shot of a bright green parrot with a red forehead and yellow wing tips, perched with plausible support on a simple wooden perch stand in a bright modern living room with a white sofa, a large window and a potted fiddle-leaf fig. Its head, beak, eyes, red forehead, green feathers, yellow wing tips, feet and perched posture are clearly visible. It remains calmly perched with natural breathing and subtle head movements. Soft indoor daylight and steady camera framing.\par\medskip

\noindent\textbf{panda\_zoo\_snowbamboo} (CLIPSeg: ``a giant panda'')\par
\noindent$\pin$: A continuous realistic medium-wide shot of a giant panda sitting on the ground inside a zoo enclosure and holding one green bamboo stalk in its front paws. A gray concrete rock wall, a wooden climbing frame and a metal fence are clearly visible behind the panda. Its face, black eye patches, ears, black-and-white fur, front paws, bamboo stalk and sitting posture are clearly visible. The panda remains calmly sitting, slowly chewing with subtle head movements. The camera framing and even daylight remain steady.\par
\noindent$\pcold$: A continuous realistic medium-wide shot of a giant panda sitting with plausible support on a snowy slope in a quiet mountain bamboo forest dusted with fresh snow, holding one green bamboo stalk in its front paws. Its face, black eye patches, ears, black-and-white fur, front paws, bamboo stalk and sitting posture are clearly visible. It remains calmly sitting, slowly chewing with subtle head movements. Soft winter daylight and steady camera framing.\par\medskip

}

%% file: supp_tables_en.tex
\begin{center}\begin{minipage}{\textwidth}
\centering\small
\captionof{table}{\textbf{Original blinded-viewer comparison, retained separately.} Counts are out of 32; clearance requires absence at 5.2 and 6.7\,s after the switch. Unknown counts are in parentheses. The man-seed-0 author adjudication affects cold restart and Molt symmetrically. Face similarity is to each condition's own pre-switch frame: the original 2.2--6.7\,s mean and the 15.7\,s value, not the 6.7\,s value of the main table. VBench is reported separately below.}
\label{tab:supp-original}
\suppfit{\begin{tabular}{lrrrrcc}
\toprule
Method & Cleared & Collage & Damage & Kept & Face 2.2--6.7\,s & Face 15.7\,s \\
\midrule
No switch (reference) & -- & -- & -- & -- & .750 & .724 \\
LongLive recache & 0 & 15 & 4 & 21 & .749 & .728 \\
Echo smooth & 0 & 11 (1) & 4 & 20 & .886 & .867 \\
Cold restart & 32 & 0 & 1 (1) & 22 & .362 & .356 \\
Echo hard cut & 32 & 0 & 4 & 20 & .390 & .396 \\
Molt Forcing & 32 & 0 & 4 & 20 & .610 & .533 \\
\bottomrule\end{tabular}}
\end{minipage}\end{center}

\begin{center}\begin{minipage}{\textwidth}
\centering\small
\setlength{\tabcolsep}{3pt}
\captionof{table}{\textbf{Original eligible subset.} 25 videos, including 15 human videos from eight families. Blinded-viewer labels with the separate man-seed-0 adjudication. Joint requires two-time clearance, kept attributes, no damage, and no collage.}
\label{tab:supp-qualified}
\suppfit{%
\begin{tabular}{lccccc}
\toprule
Method & Cleared & Collage & Kept & Joint & Face \\
\midrule
LongLive recache & 0 & 10 & 21 & 0 & 0.745 \\
Echo smooth & 0 & 9 & 19 & 0 & 0.892 \\
Cold restart & 25 & 0 & 20 & 20 & 0.366 \\
Echo hard cut & 25 & 0 & 19 & 18 & 0.381 \\
\ours & 25 & 0 & 20 & 19 & \textbf{0.616} \\
\bottomrule
\end{tabular}}
\end{minipage}\end{center}

\begin{center}\begin{minipage}{\textwidth}
\centering\small
\setlength{\tabcolsep}{3pt}
\captionof{table}{\textbf{Original short-window face similarity by human family.} Relative to each condition's own pre-switch frame, averaged over 2.2--6.7\,s after the switch and the two seeds.}
\label{tab:supp-family-arc}
\suppfit{%
\begin{tabular}{lcccc}
\toprule
Family & No switch & Cold restart & Echo hard cut & \ours \\
\midrule
businessman & 0.905 & 0.276 & 0.266 & \textbf{0.847} \\
chef & 0.775 & 0.425 & 0.465 & \textbf{0.623} \\
guitarist & 0.528 & 0.396 & 0.435 & \textbf{0.443} \\
hiker & 0.723 & 0.319 & 0.375 & \textbf{0.539} \\
man & 0.822 & 0.428 & 0.384 & \textbf{0.644} \\
painter & 0.775 & 0.277 & 0.365 & \textbf{0.635} \\
producer & 0.767 & 0.344 & 0.469 & \textbf{0.526} \\
reader & 0.828 & 0.460 & 0.451 & \textbf{0.680} \\
student & 0.868 & 0.361 & 0.474 & \textbf{0.714} \\
violinist & 0.513 & 0.335 & 0.215 & \textbf{0.446} \\
\midrule
Mean & 0.750 & 0.362 & 0.390 & \textbf{0.610} \\
\bottomrule
\end{tabular}}
\end{minipage}\end{center}

\begin{center}\begin{minipage}{\textwidth}
\centering\small
\setlength{\tabcolsep}{4pt}
\captionof{table}{\textbf{Original six VBench dimensions.} 32 videos per method; post-switch window 0--6.7\,s after the switch (108 frames), cross-switch window 2.3\,s before to 2.9\,s after (85 frames). No values are imputed for unmeasured conditions.}
\label{tab:supp-vbench}
\suppfit{%
\begin{tabular}{lcccccccccccc}
\toprule
 & \multicolumn{6}{c}{Post-switch} & \multicolumn{6}{c}{Cross-switch} \\
\cmidrule(lr){2-7}\cmidrule(lr){8-13}
Method & Subject & Background & Smooth & Dynamic & Aesthetic & Imaging & Subject & Background & Smooth & Dynamic & Aesthetic & Imaging \\
\midrule
LongLive recache & 0.995 & 0.984 & 0.996 & 0.000 & 0.631 & 0.723 & 0.996 & 0.984 & 0.996 & 0.010 & 0.634 & 0.722 \\
Echo smooth & 0.994 & 0.984 & 0.996 & 0.016 & 0.632 & 0.727 & 0.995 & 0.983 & 0.996 & 0.021 & 0.629 & 0.726 \\
Cold restart & 0.982 & 0.969 & 0.994 & 0.000 & 0.652 & 0.720 & 0.961 & 0.960 & 0.992 & 0.010 & 0.641 & 0.720 \\
Echo hard cut & 0.972 & 0.966 & 0.994 & 0.008 & 0.661 & 0.716 & 0.960 & 0.961 & 0.992 & 0.010 & 0.644 & 0.718 \\
\ours & 0.984 & 0.971 & 0.994 & 0.000 & 0.643 & 0.724 & 0.967 & 0.963 & 0.992 & 0.010 & 0.636 & 0.724 \\
\bottomrule
\end{tabular}}
\end{minipage}\end{center}

\begin{center}\begin{minipage}{\textwidth}
\centering\small
\setlength{\tabcolsep}{3pt}
\captionof{table}{\textbf{Ablation} (32 switch videos, 6.7\,s after the switch; same pre-switch videos and noise; face similarity on the 20 human videos). Cleared: all three old objects judged absent by the VLM. Collage, leakage: videos showing the artifact (unclear in parentheses). The last row changes both the window mask and recache.}
\label{tab:supp-ablation}
\suppfit{%
\begin{tabular}{lcccc}
\toprule
Configuration & Cleared$\uparrow$ & Collage$\downarrow$ & Leakage$\downarrow$ & Face$\uparrow$ \\
\midrule
LongLive recache & 0 & 4 (2) & 6 (3) & 0.724 \\
Cold restart & 31 & 0 (0) & 0 (1) & 0.350 \\
+ window reads native subject only, bridge sink & 29 & 1 (1) & 1 (4) & 0.528 \\
+ native tokens at sink subject positions & 29 & 3 (5) & 0 (7) & 0.617 \\
+ anchor mask eroded $e_2$ & 29 & 1 (2) & 1 (5) & 0.596 \\
+ outer-tightened window $o_1$ and recache (\ours) & 30 & 0 (0) & 0 (1) & 0.587 \\
\bottomrule
\end{tabular}}
\par\vspace{2pt}
\end{minipage}\end{center}

\begin{center}\begin{minipage}{\columnwidth}
\centering\small
\setlength{\tabcolsep}{3.5pt}
\captionof{table}{\textbf{Window $\times$ sink $2\times2$ replacement} (32 switch videos, 6.7\,s after the switch). Cleared: all three old objects judged absent by the VLM. Collage: spliced old and new scene regions; unclear in parentheses. Face: similarity to the pre-switch frame on the 10 human families. No configuration recaches.}
\label{tab:diag}
\suppfit{%
\begin{tabular}{llcccc}
\toprule
Window & Sink & Cleared$\uparrow$ & Collage$\downarrow$ & Bg.\ progress$\uparrow$ & Face$\uparrow$ \\
\midrule
native & native & 0/32 & 5 (1) & 0.07 & 0.734 \\
native & cold & 3/32 & 22 (2) & 0.56 & 0.402 \\
cold & native & 20/32 & 15 (4) & 0.82 & 0.363 \\
cold & cold & 31/32 & 0 (0) & 0.99 & 0.350 \\
\bottomrule
\end{tabular}}
\par\vspace{2pt}
\end{minipage}\end{center}

%% file: supp_cost_en.tex
\begin{center}\begin{minipage}{\textwidth}
\centering\small
\setlength{\tabcolsep}{4pt}
\captionof{table}{\textbf{Original post-switch generator calls.} Human videos: 21 blocks (15.75\,s after the switch); non-human: nine blocks (6.75\,s).}
\label{tab:supp-cost}
\suppfit{%
\begin{tabular}{lccc}
\toprule
Method & Extra calls & Human & Non-human \\
\midrule
LongLive recache & recache 1 & 106 & 46 \\
Echo smooth & -- & 105 & 45 \\
Echo hard cut & -- & 105 & 45 \\
Cold restart & Cold block 5 & 110 & 50 \\
\ours & recache 1 + Cold block 5 & 111 & 51 \\
\bottomrule
\end{tabular}}
\end{minipage}\end{center}

%% file: supp_followup_en.tex
\paragraph{Scope and timing of predictions.}
The initial follow-up prediction document and its first supplement were frozen before any follow-up inference.
A later scope reduction preceded all follow-up video generation but followed a text-only rewriting stage: 30/32 proposed automatic target prompts did not meet its strict format, so the automatic-prompt group was dropped.
Seven predictions are retained, plus the nondirectional long-horizon identity readout.
Registered checks on RoPE cut, no recache, pixel composition, an Echo no-switch run, automatic prompts, and anchor removal in this follow-up matrix were not run as video tests.
The RoPE slope check retains only switch-condition horizontal shifts; stay and vertical-shift controls and the 120\,s extension were removed.
Table~\ref{tab:predictions} reports every retained prediction.

\begin{center}\begin{minipage}{\textwidth}
\centering\small
\captionof{table}{\textbf{Retained prospective checks.} Thresholds preceded follow-up video generation. Statistical implementation was frozen later, before aggregation.}
\label{tab:predictions}
\begin{tabular}{@{}p{2.3cm}p{5.2cm}p{8.4cm}@{}}
\toprule
Check & Registered readout & Observation and interpretation \\
\midrule
KV flush & Clearance $<20/32$ and original-definition DINO progress below window replacement & 0/32; window replacement minus KV flush .743 [.668, .812]. Both parts satisfied. \\
Sink recomputation & Clearance $\le3/32$; $>6/32$ explicitly contradicts & 0/32. Threshold satisfied. \\
RoPE slope & Horizontal slope 12--20 px/token; absolute slope below 4 explicitly contradicts & Outside means 12.27--14.04 fall in the range; inside means 4.89--6.45 have CIs including zero. \\
Attention concentration & At least 4/7 sampled layers exceed ten times uniform local-attention expectation & 7/7 using actual visible-key normalization; also 7/7 under the original approximate normalization. Point threshold satisfied. \\
Residue at 60\,s & At most 2/20 Molt videos have any old object present 60\,s after the switch & 0/20, unclear 0; Echo also 0/20. \\
Slot shares & Window-only displacement at least 75\% of all-slot displacement; sink-only at most 25\% & Both regional means satisfy the constraints; the window contrast CIs include zero. \\
Anchor attention & Subject query attention to anchor at least twice background query attention in at least 4/7 layers & 6/7 layers meet the point threshold. \\
Identity at 60\,s & Write ``advantage at 60 s'' only if the paired face-similarity CI lower bound at 60\,s is $>0$ & .113 [$-.004$, .234]. Condition not met; the paper claims the advantage only through about 16\,s. \\
\bottomrule\end{tabular}
\end{minipage}\end{center}

\paragraph{Spatial measurements.}
Eight prespecified seed-0 videos are the resampling units; attention heads, layers, steps, and blocks are not independent videos.
Only pre-switch historical keys receive the phase change; newly generated keys do not accumulate it.
All six horizontal/vertical slope strata are retained in Table~\ref{tab:g4-full}.
The prediction specified no vertical threshold.

\begin{center}\begin{minipage}{\textwidth}
\centering\small
\captionof{table}{\textbf{Spatial RoPE slopes} (pixels/token, video-bootstrap 95\% CI). Vertical response is measured under horizontal key shifts; no vertical-shift control was run.}
\label{tab:g4-full}
\begin{tabular}{llcc}
\toprule
Region & Time after switch & Horizontal & Vertical \\
\midrule
Inside subject box & 0.7\,s & 4.887 [$-3.217$, 13.304] & .014 [.005, .023] \\
 & 1.4\,s & 5.674 [$-1.152$, 13.336] & $-.040$ [$-.123$, .021] \\
 & 2.2\,s & 6.454 [$-.707$, 13.376] & .046 [$-.017$, .150] \\
Outside subject box & 0.7\,s & 14.043 [7.686, 18.437] & $-.596$ [$-1.347$, $-.015$] \\
 & 1.4\,s & 13.177 [7.217, 16.938] & $-.385$ [$-.921$, $-.061$] \\
 & 2.2\,s & 12.272 [6.669, 16.172] & $-.409$ [$-1.783$, .664] \\
\bottomrule\end{tabular}
\end{minipage}\end{center}

\begin{center}\begin{minipage}{\textwidth}
\centering\small
\captionof{table}{\textbf{Signed displacement 0.7\,s after the switch under an eight-token shift.} A: all pre-switch history; W: window only; S: sink only. Means and 95\% CIs, pixels. Proportions are undefined when the all-slot mean is nonpositive; signed contrasts are retained.}
\label{tab:g4-slot}
\suppfit{\begin{tabular}{lccccc}
\toprule
Region & A & W & S & W$-0.75$A & $0.25$A$-$S \\
\midrule
Inside & 16.62 [$-64.20$, 100.75] & 36.45 [$-36.23$, 103.11] & 1.66 [.84, 2.54] & 23.98 [$-3.76$, 59.01] & 2.49 [$-17.46$, 23.89] \\
Outside & 104.94 [9.96, 177.01] & 86.31 [$-2.88$, 158.09] & .91 [.16, 1.92] & 7.60 [$-29.34$, 37.15] & 25.32 [1.71, 43.00] \\
\bottomrule\end{tabular}}
\end{minipage}\end{center}

\paragraph{Long-horizon heterogeneity.}
At 60\,s after the switch, Molt, Echo, and no-switch similarities to the pre-switch frame are .496 [.403, .593], .384 [.322, .439], and .613 [.480, .717].
The paired Molt--Echo sign test is 8/2/0 positive/negative/tied families, two-sided $p=.109$.
The first sampled difference interval whose lower bound is not positive occurs at 45.7\,s.
Table~\ref{tab:long-families} retains all families, including negative differences.
Measured against each output's own frame at 0.7\,s, the paired difference at 60\,s is $-.204$ [$-.305$, $-.118$]: Molt stays closer to the pre-switch subject, and Echo stays closer to its own post-switch output.

\begin{center}\begin{minipage}{\textwidth}
\centering\small
\captionof{table}{\textbf{Molt--Echo face-similarity difference 60\,s after the switch}, each relative to its own pre-switch frame. Two seeds and their family mean, all ten human families.}
\label{tab:long-families}
\begin{tabular}{lrrr}
\toprule
Family & Seed 0 & Seed 1 & Mean \\
\midrule
Businessman & .630 & .405 & .518 \\
Chef & .077 & $-.028$ & .024 \\
Guitarist & $-.161$ & $-.365$ & $-.263$ \\
Hiker & .010 & .330 & .170 \\
Man & .228 & .107 & .168 \\
Painter & .083 & .188 & .136 \\
Producer & $-.231$ & .076 & $-.078$ \\
Reader & .150 & .117 & .133 \\
Student & .269 & .114 & .192 \\
Violinist & .150 & .109 & .129 \\
\bottomrule\end{tabular}
\end{minipage}\end{center}

\paragraph{CLIP and background proxies.}
Table~\ref{tab:clip-full} reports all eight conditions with the common target prompt.
The Molt--Echo all-video CLIP difference is $-.004$ [$-.008$, .002], and the family sign test gives 3 positive and 13 negative families ($p=.021$).
The interval measures the mean difference, and the sign test counts the direction per family.
Molt minus KV flush and Molt minus sink recomputation are .063 [.052, .076] and .066 [.052, .081], respectively.
Table~\ref{tab:dino-full} reports both background definitions.

\begin{center}\begin{minipage}{\textwidth}
\centering\small
\captionof{table}{\textbf{CLIP text cosine 6.7\,s after the switch, family means and pointwise 95\% CIs.} All conditions use the same manually written target prompt.}
\label{tab:clip-full}
\begin{tabular}{lccc}
\toprule
Condition & All 32 videos (16 families) & 20 human (10 families) & 12 non-human (6 families) \\
\midrule
Native history & .315 [.300, .331] & .315 [.294, .338] & .315 [.302, .327] \\
Replace sink & .363 [.349, .377] & .365 [.343, .386] & .361 [.347, .375] \\
Replace window & .374 [.363, .386] & .380 [.364, .397] & .365 [.356, .375] \\
Cold restart & .380 [.369, .391] & .384 [.368, .400] & .372 [.364, .383] \\
Molt & .378 [.367, .390] & .382 [.365, .399] & .372 [.361, .385] \\
Echo hard cut & .382 [.371, .392] & .382 [.367, .397] & .381 [.370, .396] \\
KV flush & .315 [.300, .329] & .313 [.292, .334] & .319 [.303, .334] \\
Sink recomputation & .313 [.297, .329] & .312 [.289, .335] & .314 [.297, .329] \\
\bottomrule\end{tabular}\end{minipage}\end{center}

\begin{center}\begin{minipage}{\textwidth}
\centering\small
\captionof{table}{\textbf{Background progress 6.7\,s after the switch, all 32 videos.} Original and source-to-cold-normalized definitions from Sec.~F.}
\label{tab:dino-full}
\begin{tabular}{lcc}
\toprule
Condition & Original progress & Added normalized progress \\
\midrule
Native history & .067 [.019, .134] & .064 [.019, .128] \\
Replace sink & .560 [.441, .678] & .522 [.405, .635] \\
Replace window & .820 [.763, .876] & .855 [.789, .917] \\
Cold restart & .987 [.984, .990] & .998 [.994, 1.003] \\
Molt & .966 [.960, .972] & .989 [.980, .997] \\
Echo hard cut & .917 [.899, .933] & .997 [.980, 1.012] \\
KV flush & .078 [.024, .150] & .073 [.023, .141] \\
Sink recomputation & .065 [.016, .134] & .061 [.016, .125] \\
\bottomrule\end{tabular}\end{minipage}\end{center}

\paragraph{Motion experiment.}
A first stage tests 48 family--action pairs (small/medium/large in 16 families), each with three cold-start samples and no-action controls.
Selection requires action completion in at least 2/3 samples and at most 1/3 no-action controls; 15 pairs qualify (5 small, 8 medium, 2 large).
The second stage uses two seeds and five conditions, 150 new videos, with original no-action clips reused for separate questions.
Action labels come from an uncalibrated nine-frame VLM proxy; all 270 second-stage predictions give a category but omit the requested explanation.
Static review covered 22 prespecified videos at 198 sampled frames.

\begin{center}\begin{minipage}{\textwidth}
\centering\small
\captionof{table}{\textbf{Action completion counts.} Large motion contains only canine pairs. ``Molt, no anchor'' removes the persistent anchor from the second block after the switch.}
\label{tab:motion}
\begin{tabular}{lrrrrr}
\toprule
Stratum & Cold restart & Molt & Molt, no anchor & Echo hard cut & No switch \\
\midrule
Small (/10) & 2 & 4 & 4 & 4 & 7 \\
Medium (/16) & 9 & 10 & 10 & 11 & 12 \\
Large (/4) & 1 & 0 & 0 & 0 & 1 \\
All (/30) & 12 & 14 & 14 & 15 & 20 \\
\bottomrule\end{tabular}\end{minipage}\end{center}

Cold restart completes while Molt does not in two pairs; the reverse occurs in four.
Removing the persistent anchor does not recover completion in those two pairs.
The subset in which cold restart or no switch completes contains 21 pairs and retains the same two/four discordance; this subset is selected on control outcomes.
The mask-based motion measurements mix pose, viewpoint, scale, occlusion, and segmentation changes.

\paragraph{Author continuous-video subset.}
The author submitted 47 complete, personally attested forms: all 24 long-horizon items (four human families, two seeds, Molt/Echo/no switch), 21/56 spatial-intervention items, and the two businessman-seed-0 KV flush and sink recomputation items.
The other 35 spatial forms were left incomplete.
The author's additional group-level statement reports persistent old objects and generally preserved human appearance in the spatial-intervention group, without per-item labels.
The software playback-percentage gate was removed at the author's request; the attestations are self-reports.
The Molt long-horizon subset is described as no observed old objects and similar identity, with the two chef facial-damage exceptions in Sec.~H.
This subset is non-blind and covers four families.

\paragraph{Appearance details beyond the face.}
At 6.7\,s after the switch, Echo hard cut redraws clothing details that the target prompt leaves unspecified (\cref{fig:supp-appearance}).
In both violinist seeds the sleeveless dress becomes long-sleeved; in both hiker seeds the backpack chest strap disappears and the zipper changes color.
\ours keeps these details in all four videos.
Over the following minute, Echo's regenerated appearance stays nearly fixed, whereas \ours drifts gradually, for example the hiker's jacket darkens in seed 0 and the guitarist's inner shirt changes from red to white in seed 1.
One account is that Echo re-synthesizes subject and scene together and then continues its own output, whereas \ours keeps reconciling native subject tokens with context increasingly generated in the new scene.
This account is untested; Echo's lower similarity at the cut, its different cache configuration, and backbone drift that also appears without a switch may also contribute to the narrowing identity gap.
These observations come from unblinded inspection of the 20 human videos at 6.7\,s and the eight long-horizon videos to 60\,s; they are not quantified.
The prompt-level keep-attribute checks list items only at the category level (e.g., ``long burgundy dress'') and mostly mark these cases as kept or unknown.

\begin{center}\begin{minipage}{.82\linewidth}
\centering\includegraphics[width=\linewidth]{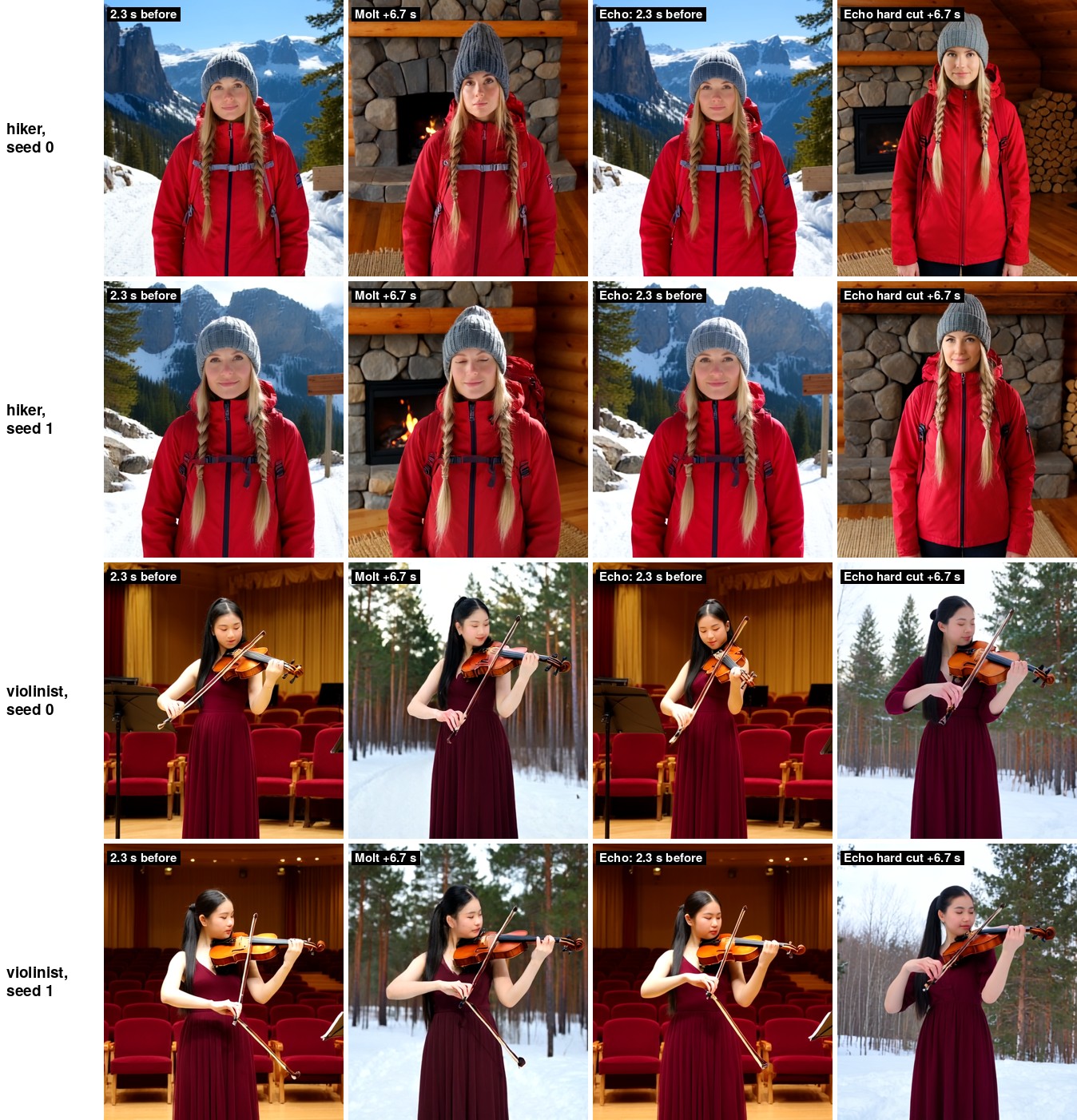}
\captionof{figure}{\textbf{Clothing details at the switch.} Central half-width crops of hiker and violinist, seeds 0 and 1. Columns: LongLive frame 2.3\,s before the switch, \ours at +6.7\,s, Echo's own frame 2.3\,s before the switch, and Echo hard cut at +6.7\,s. The two systems have different pre-switch videos, so each is paired with its own.}
\label{fig:supp-appearance}\end{minipage}\end{center}

\paragraph{Additional limits.}
False application of the method to same-scene edits rearranges layout in the tested cases (background jump .097 versus .0004 for recache).
A text trigger separates explicit location changes well, but a frozen threshold transfers poorly to harder prompts (recall .067 despite AUC .993).
The tested gradual-transition prompts and spatial handoffs do not establish clean gradual in-shot transitions.
Multiple switches, returns, and generalization to native configurations of other backbones remain unvalidated.